\documentclass[acmtog]{acmart}
\usepackage{booktabs} % For formal tables
\usepackage{bm}
\usepackage{subcaption}
\usepackage{enumitem}
\usepackage[ruled]{algorithm2e} % For algorithms
\usepackage{algpseudocode}

\usepackage{hyperref}
\begin{document}
	% Title portion
	\title{ColorizeDiffusion: Adjustable Sketch Colorization with Reference Image and Text}
	
	% DO NOT ENTER AUTHOR INFORMATION FOR ANONYMOUS TECHNICAL PAPER SUBMISSIONS TO SIGGRAPH 2019!
	\author{Dingkun Yan}
	\affiliation{%
		  \institution{Tokyo Institute of Technology}
		  \department{School of Computing}
		  \state{Tokyo}
		  \country{Japan}}
	\email{yan@img.cs.titech.ac.jp}
 
	\author{Liang Yuan}
	\affiliation{%
		  \institution{Keio University}
		  \state{Kanagawa}
		  \country{Japan}
		}
  
  	\author{Erwin Wu}
        \affiliation{%
            \institution{Tokyo Institute of Technology}
    	\department{School of Computing}
    	\state{Tokyo}
	   	\country{Japan}
        }

	\author{Yuma Nishioka}
	\affiliation{%
    	\institution{Tokyo Institute of Technology}
    	\department{School of Computing}
    	\state{Tokyo}
            \country{Japan}
        }
  
	\author{Issei Fujishiro}
	\affiliation{%
		  \institution{Keio University}
		  \state{Kanagawa}
		  \country{Japan}
		}
 
	\author{Suguru Saito}
	\affiliation{%
    	\institution{Tokyo Institute of Technology}
    	\department{School of Computing}
    	\state{Tokyo}
    	\country{Japan}
        }
	%\author{Tian He}
	%\affiliation{%
		%  \institution{University of Virginia}
		%  \department{School of Engineering}
		%  \city{Charlottesville}
		%  \state{VA}
		%  \postcode{22903}
		%  \country{USA}
		%}
	%\affiliation{%
		%  \institution{University of Minnesota}
	%\email{tinghe@uva.edu}
	%\author{Chengdu Huang}
	%\author{John A. Stankovic}
	%\author{Tarek F. Abdelzaher}
	%\affiliation{%
		%  \institution{University of Virginia}
		%  \department{School of Engineering}
		%  \city{Charlottesville}
		%  \state{VA}
		%  \postcode{22903}
		%  \country{USA}
		%}
	
	%\renewcommand\shortauthors{Zhou, G. et al}
	\begin{teaserfigure}
		\centering
		\includegraphics[width=1\linewidth]{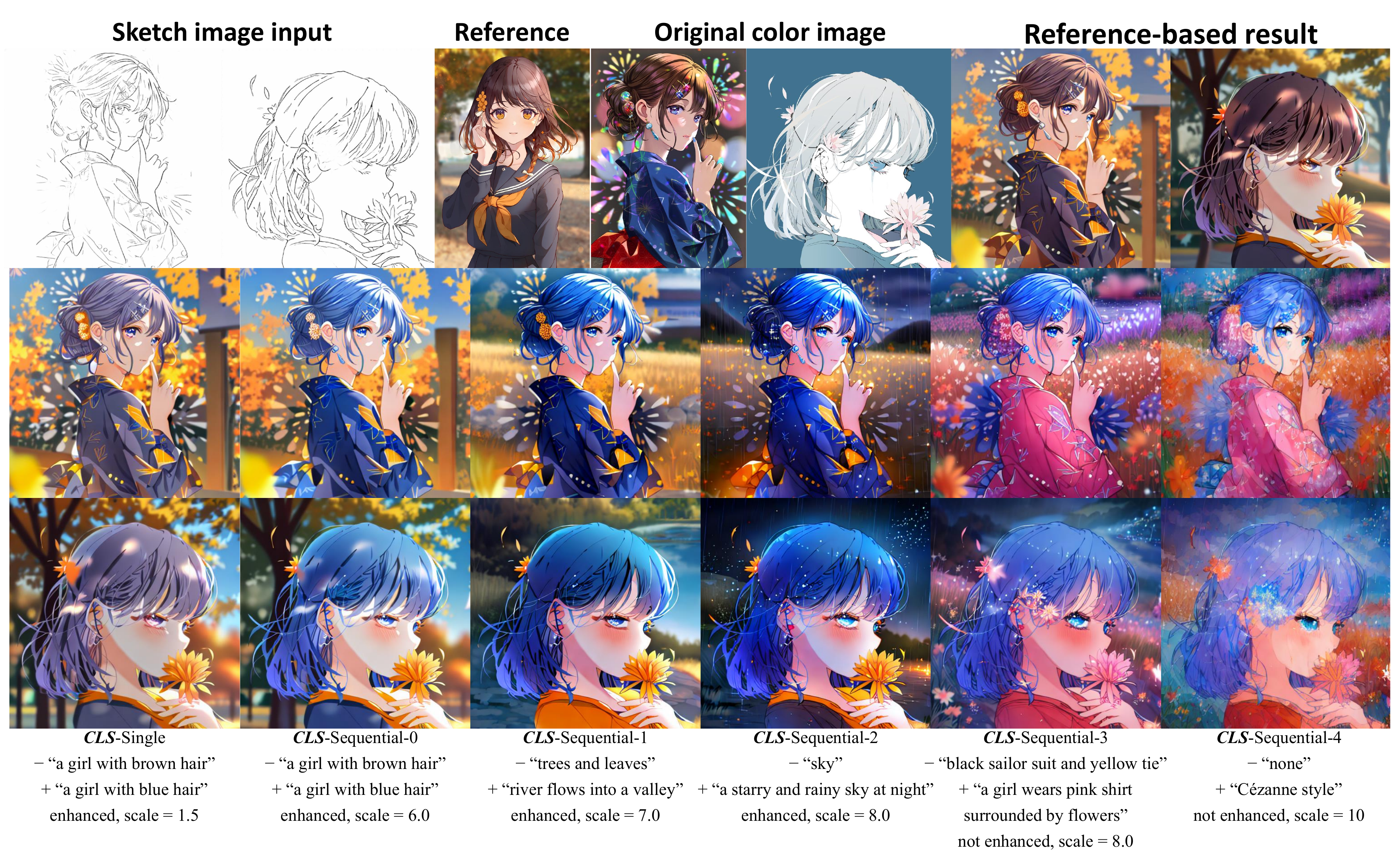}
		\caption{Our method colorizes sketch images based on a reference image and allows the results to be sequentially edited using arbitrary text inputs with specified degrees. Symbols ``+'' and ``--'' respectively denote the target text and anchor text for our text-based manipulation.}
		\label{teaserfigure}
	\end{teaserfigure}
	\begin{abstract}
            Diffusion models have recently demonstrated their effectiveness in generating extremely high-quality images and are now utilized in a wide range of applications, including automatic sketch colorization. Although many methods have been developed for guided sketch colorization, there has been limited exploration of the potential conflicts between image prompts and sketch inputs, which can lead to severe deterioration in the results. Therefore, this paper exhaustively investigates reference-based sketch colorization models that aim to colorize sketch images using reference color images. We specifically investigate two critical aspects of reference-based diffusion models: the ``distribution problem'', which is a major shortcoming compared to text-based counterparts, and the capability in zero-shot sequential text-based manipulation. We introduce two variations of an image-guided latent diffusion model utilizing different image tokens from the pre-trained CLIP image encoder and propose corresponding manipulation methods to adjust their results sequentially using weighted text inputs. We conduct comprehensive evaluations of our models through qualitative and quantitative experiments as well as a user study.
	\end{abstract}
	
	\begin{CCSXML}
		<ccs2012>
		<concept>
		<concept_id>10010405.10010469.10010470</concept_id>
		<concept_desc>Applied computing~Fine arts</concept_desc>
		<concept_significance>500</concept_significance>
		</concept>
		<concept>
		<concept_id>10010147.10010178.10010224</concept_id>
		<concept_desc>Computing methodologies~Computer vision</concept_desc>
		<concept_significance>500</concept_significance>
		</concept>
		<concept>
		<concept_id>10010147.10010371.10010382.10010383</concept_id>
		<concept_desc>Computing methodologies~Image processing</concept_desc>
		<concept_significance>300</concept_significance>
		</concept>
		</ccs2012>
	\end{CCSXML}
	
	\ccsdesc[500]{Applied computing~Fine arts}
	\ccsdesc[500]{Computing methodologies~Computer vision}
	\ccsdesc[300]{Computing methodologies~Image processing}
	
	\keywords{Sketch colorization, Dual-conditioned generation, Latent diffusion model, Latent manipulation}
	
	\maketitle
	
	\section{Introduction}
\indent Anime-style images have gained worldwide popularity over the past few decades thanks to their diverse color composition and captivating character design, but the process of colorizing sketch images has remained labor-intensive and time-consuming. However, swift advancements to diffusion models \cite{HoJA20, controlnet} now enable large generative models to create remarkably high-quality images across various domains, including anime style. Most conditional diffusion models predominantly focus on text-based generation, and few specialize in the reason for the deterioration when applying image-guided models to reference-based sketch colorization, a complex dual-conditioned generation task that utilizes both a reference and a sketch image. As such, this paper focuses on reference-based colorization by thoroughly analyzing this reason for deterioration, which is the major challenge in training-related models. We explore training strategies for relevant neural networks and propose two zero-shot text-based manipulation methods using tokens from pre-trained CLIP encoders.\\
\indent A salient issue in the multi-conditioned generation is the potential conflict between input conditions. While this might not significantly impact methods using sketch and text conditions, such conflicts are problematic in reference-based colorization because both sketch and reference images contain varied information about structure, location, and object identity, with potentially incompatible contents. This issue, termed the ``distribution problem'' in this paper, stems from the semantic alignment of training data, where reference images used in training always correspond to the ground truth, and the networks accordingly prioritize reference embeddings over sketch semantics during inference. We investigate three feasible methods for addressing this issue and consider the most effective solution to be the one that adds timestep-dependent noise to the reference embeddings during training. The investigation of and solution to the distribution problem constitute the key points of this paper.\\
\indent Text-based models, despite their advantages, also have several limitations in comparison to image-guided methods. Two notable limitations are their inability to accurately transfer features from reference images and to effectively reflect the progressive changes in results due to weighted text inputs \cite{RombachBLEO22,ruiz2022dreambooth,HuSWALWWC22}, a process often referred to as ``latent interpolation'' \cite{abs-2204-06125}. When trained using image features that adapt in response to the confidence of corresponding attributes, image-guided models \cite{abs-2204-06125,PatashnikWSCL21,KimKY22a,GalPMBCC22,LiuPAZCHSRD23,ip-adapter} have shown potential to effectively address this issue with zero-shot algorithms.\\
\indent Given that anime-style images \cite{danbooru2021} are more sensitive to color variations and encapsulate ample visual attributes within each image, they are suitable to aid in analyzing the proposed reference-based generation and text-based manipulation methods. Our research demonstrates that reference-based models, leveraging image tokens from pre-trained CLIP encoders as conditions, are capable of progressively adapting their outputs in response to weighted text inputs.\\
\indent Through rigorous experimentation with ablation models and baselines, we empirically prove the effectiveness of the proposed methods in reference-based colorization and text-based manipulation. We further conducted a user study to evaluate the proposed methods subjectively.\\
\indent The contributions of this paper can be summarized as follows:
\begin{itemize}[leftmargin=*]
    \item[$\bullet$] We conduct a comprehensive investigation of the distribution problem in reference-based sketch colorization training using latent diffusion models. To better explore this problem, we propose various reference-based models.
    \item[$\bullet$] We offer a general solution to diminish the distribution problem discussed in this paper.
    \item[$\bullet$] We design two zero-shot manipulation methods for reference-based models using different types of image tokens.
\end{itemize}

	\section{Related Work}
	Our work focuses on reference-based sketch colorization, an important subfield of image generation. We utilize the score-based generative model \cite{HoJA20,0011SKKEP21,RombachBLEO22} as our neural backbone, which is widely known as the diffusion model. Our training methods and overall pipeline are designed following previous style transfer and colorization methods, pursuing pixel-level correspondence and fidelity to the input sketch image.\\
	
	\noindent\textbf{Latent Diffusion Models.} Diffusion probabilistic Models (DMs) \cite{HoJA20} are a class of latent variable models inspired by considerations from nonequilibrium thermodynamics \cite{Sohl-DicksteinW15}. Compared with Generative Adversarial Nets (GANs) \cite{GoodfellowPMXWOCB14,KarrasLA19,KarrasLAHLA20,ChoiCKH0C18,ChoiUYH20}, DMs excel at generating highly realistic images across various contexts. However, the autoregressive denoising process, typically computed using a deep U-Net network \cite{RonnebergerFB15}, incurs substantial computational costs for both training and inference, which limits further applications. To address this limitation, LDM \cite{RombachBLEO22}, also known as StableDiffusion (SD) and SDXL \cite{sdxl}, utilizes a two-stage synthesis and carries out the diffusion/denoising process within a highly compressed latent space to reduce computational costs significantly. Concurrently, many efficient samplers have been proposed to accelerate the denoising process \cite{SongME21,0011SKKEP21,0011ZB0L022,abs-2211-01095}. In this paper, we adopt a pre-trained text-based SD model as our neural backbone, utilize DPM++ solver and Karras noise scheduler \cite{abs-2211-01095,0011SKKEP21,KarrasAAL22} as the default sampler, and employ classifier-free guidance \cite{DhariwalN21,abs-2207-12598} to strengthen the reference-based performance.\\
	
	\noindent\textbf{Neural Style Transfer.} First proposed in \cite{GatysEB16}, Neural Style Transfer (NST) has now become a widely adopted technique compatible with many effective generative models. Reference-based colorization, which aims to transfer colors and textures from reference images to sketch images, can be viewed as a subclass of multi-domain style transfer. However, compared to traditional network-based NST methods \cite{JohnsonAF16,HuangB17,ZhuPIE17,ChoiCKH0C18,ChoiUYH20}, which typically train networks using feature-level restrictions, reference-based colorization requires a higher level of color correspondence with the reference while maintaining fidelity to the sketch inputs. Consequently, our method is developed based on the principles of conditional image-to-image translation \cite{IsolaZZE17} to ensure pixel-level correspondence between the sketch and colorized results. We also demonstrate the efficiency of our approach to sketch-based style transfer.\\

	\noindent\textbf{Image Colorization.} Developing automatic colorization algorithms has been a popular topic in the image generation field for years. Many effective methods have been developed for this purpose, all of which can be divided into traditional \cite{SykoraDC09,cgf.14517,FurusawaHOO17,RevoyFT18} or Deep Learning (DL)-based methods \cite{ZhangIE16,IsolaZZE17,he2018deep} according to the adoption of deep neural networks. Our work is highly related to DL-based methods, as they have proven effective in generating high-quality images and controlling outputs using various conditional inputs. According to the conditions, existing DL-based methods can be categorized into three types: text-based \cite{zouSA2019sketchcolorization,controlnet,KimJPY19}, user-guided \cite{ZhangZIGLYE17,ZhangLW0L18}, and reference-based \cite{SunLWW19,LeeKLKCC20,AkitaMT20,yan-cgf}. Text-based methods adopt text tags/prompts as hints to guide colorization, and they are the most popular subclass nowadays, owing to sufficient pre-trained Text-to-Image (T2I) models, as well as many practical plug-in modules and fine-tuning methods \cite{controlnet,HuSWALWWC22,ruiz2022dreambooth}. However, most text-based models cannot precisely adjust the scale of specific prompts or transfer features from references without training, while user-guided methods require users to specify colors manually for each region using color spots or spray \cite{ZhangLW0L18}, assuming the user has a basic knowledge of line art. Yan et al. investigated the possibility of combining image and text tag conditions \cite{yan-cgf}, but it was ineffective at generating backgrounds and at handling complex references, like many other GAN-based methods \cite{ChoiUYH20,LeeKLKCC20,li2022eliminating}. To overcome the limitations of reference-based methods, we comprehensively investigate the application of image-guided LDMs and propose novel manipulation methods to enable text-based control.
	\begin{figure}
    \centering
    \includegraphics[width=1\linewidth]{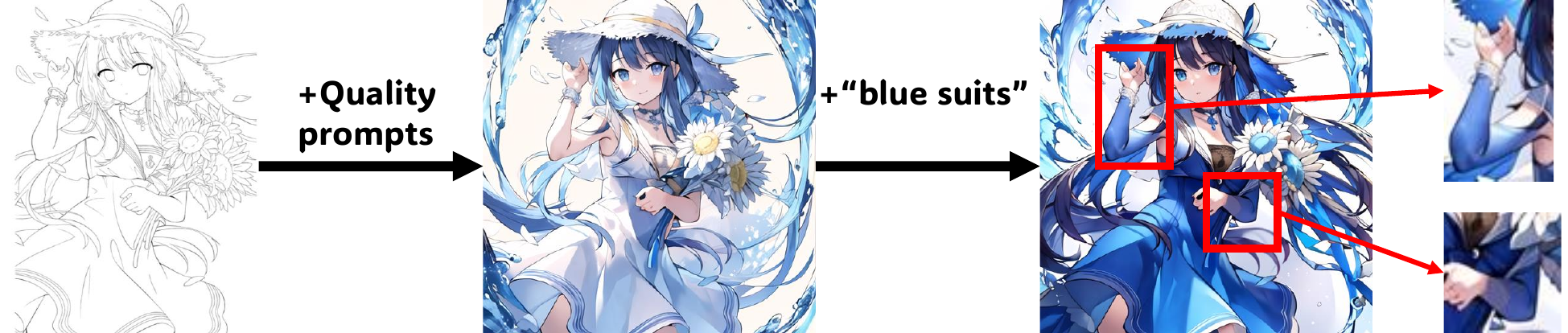}
    \caption{Illustration of distribution problem in T2I colorization. The network prioritizes prompt conditions over the sketch in the arm regions. This preference results in unexpected colorization discrepancies, particularly in areas anticipated to be skin-toned, thereby leading to visually discordant segmentation. Presented results are derived from the \textit{ControlNet\_lineart\_anime + Anything v3} framework.}
    \label{t2i-example}
\end{figure}
\section{Reference-based colorization}
In this section, we briefly outline the workflow of LDMs in Section 3.1 and present the formulation of the so-called ``distribution problem'' that arises when applying LDMs to reference-based sketch colorization in Section 3.2. We propose various training strategies to tackle the distribution problem in Sections 3.3 and 3.4.
\begin{figure}[t]
	\centering
		\includegraphics[width=1\linewidth]{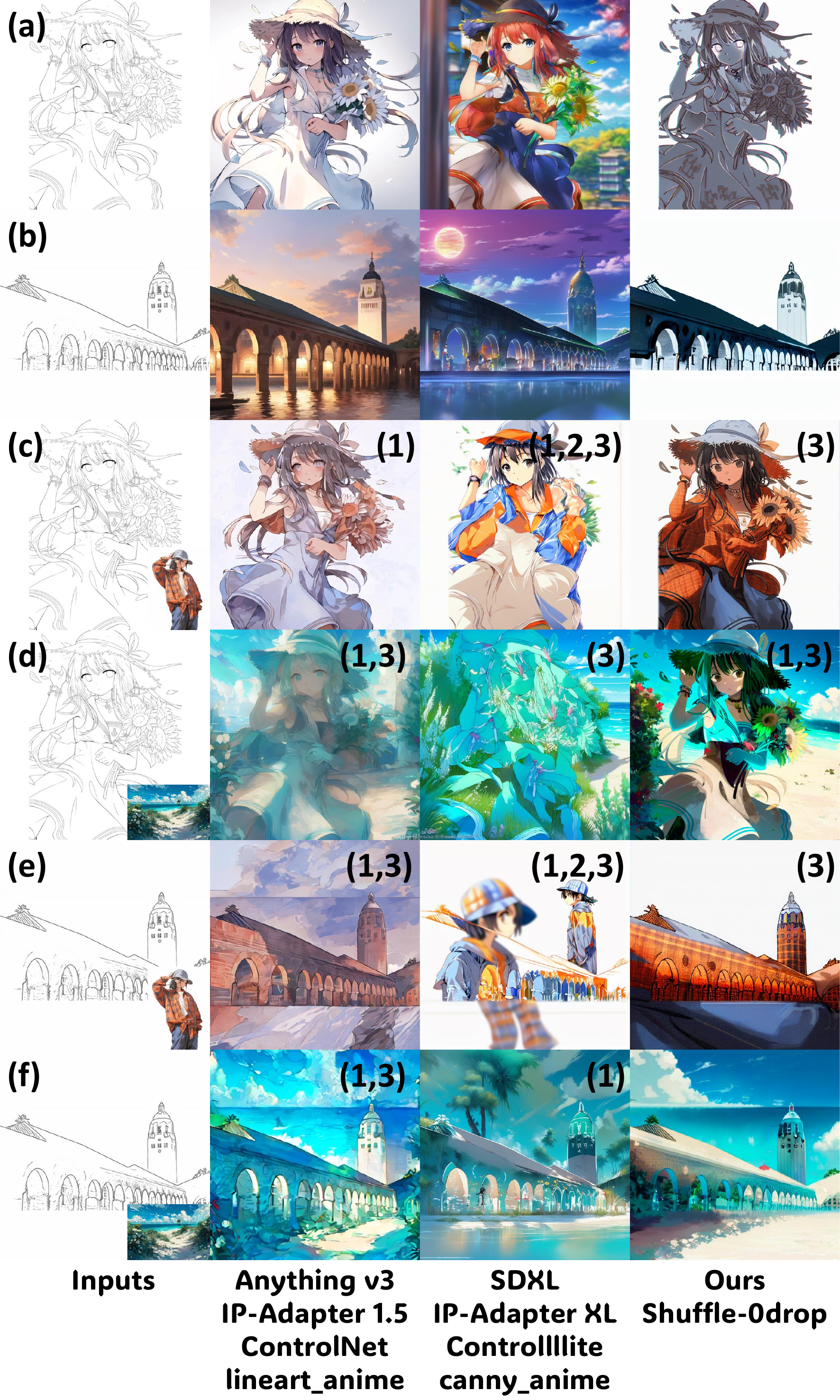}
	\caption{Illustration of deterioration caused by the distribution problem: (1) quality of textures, (2) erroneously rendered objects, and (3) segmentation error. \textit{Shuffle-0drop} is one of our ablation models.}
	\label{distribution-vis}
\end{figure}

\subsection{Latent Diffusion and Denoising}
\begin{itemize}[leftmargin=*]
	\item[1.] Train a Variational AutoEncoder (VAE) \cite{KingmaW13} on the target image domain, comprising an encoder $\mathcal{E}$ and a decoder $\mathcal{D}$ for perceptual compression and decompression, respectively.
	\item[2.] The encoder $\mathcal{E}$ compresses an image $y$ into latent representations $z_{0}=\mathcal{E}(y)$ based on a scaling factor $f$, which is defined as $f=\frac{H}{h}=\frac{W}{w}$, where ($H,W$) and ($h,w$) denote the (height, width) of the input image and the latent representations, respectively. We set the scaling factor to 8 following popular SD models.
	\item[3.] Autoregressively add noise $\epsilon\sim\mathcal{N}(0,1)$ to $z_{0}$ through $z_{t}=\alpha_{t}z_{0}+\beta_{t}\epsilon$, where $t$ denotes the timestep, $z_{t}$ the noisy representations, and $\alpha_{t}$ and $\beta_{t}$ the hyper-parameters that control the schedule of added noise. This process, known as ``diffusion'', is a fixed-length Markovian process with $T$ steps in total, where $T$ is set to $1,000$ in practice. The denoising U-Net $\theta$ learns to predict the noise $\epsilon$ at the $t$-step using the following function:
	\begin{equation}
		\mathcal{L}(\theta)=\mathbb{E}_{\mathcal{E}(y),\epsilon,t,c}[\|\epsilon-\epsilon_{\theta}(z_{t},t,c)\|^{2}_{2}],
		\label{LatentDiffLoss}
	\end{equation}
	where $c$ denotes the guiding condition.
	\item[4.] The denoising U-Net predicts $\epsilon_{t}$ to denoise $z'_{T}$ to $z'_{0}$ autoregressively during the inference stage, where $z'$ is the generated representation and $z'_{T}$ is usually a random noise sampled from a normal distribution. 
	\item[5.] Decompress the final latent representation to obtain the final image output $y'$ using the decoder $\mathcal{D}$, expressed as $y'=\mathcal{D}(z'_{0})$.
\end{itemize}
Note that only steps 4 and 5 are undertaken during inference.

\subsection{Distribution Problem}
We introduce a significant challenge in image-guided colorization, termed the ``distribution problem'', which is an issue often mistakenly identified as a type of recognition error. An example of the distribution problem in T2I colorization is given in Figure \ref{t2i-example}. Unlike text- or user-guided colorization, where conflicting conditions are less likely to arise during inference, image-guided methods often involve spatial information in the reference embeddings. This spatial information can become entangled with the forward features inside the denoising model, leading to a severe deterioration in the quality of generated images. As illustrated in Figure  \ref{distribution-vis}, networks whose adapters are trained independently generally produce inferior results compared to those generated using the respective condition independently. To facilitate understanding, we explain this problem from three different perspectives, as follows.

1. The spatial information inside the reference embeddings becomes entangled with the forward features. As previously stated, the reference embeddings used in image-guided models usually involve spatial information, more or less, depending on their preprocessing and dropping. In contrast to other dual-conditioned generations, sketch colorization should prioritize sketch semantics over reference conditions. Therefore, visually unpleasant segmentations of the \textit{Shuffle-0drop} model can be observed in Figure \ref{distribution-vis}, since it prioritizes the reference embeddings rather than the sketch semantics.

2. The DM tends to degrade into a decoder of the pre-trained encoder. While making a generative model, the decoder of a pre-trained encoder is the target of many types of generation, which is not desirable in image-guided colorization. Compared to GANs, DMs exhibit significantly better generation ability, as they are capable of reconstructing images using even only the CLS token from a pre-trained ViT \cite{ip-adapter,openclip}. However, in such cases, sketch images become less meaningful for the models, and they are likely to overlook the semantics provided by sketch inputs. Although training the entire network using the CLS token improves the prioritization of spatial information from sketches, this method becomes less efficient when local tokens are utilized to enhance resemblance with reference images.
\begin{figure}[t]
	\centering
		\includegraphics[width=1\linewidth]{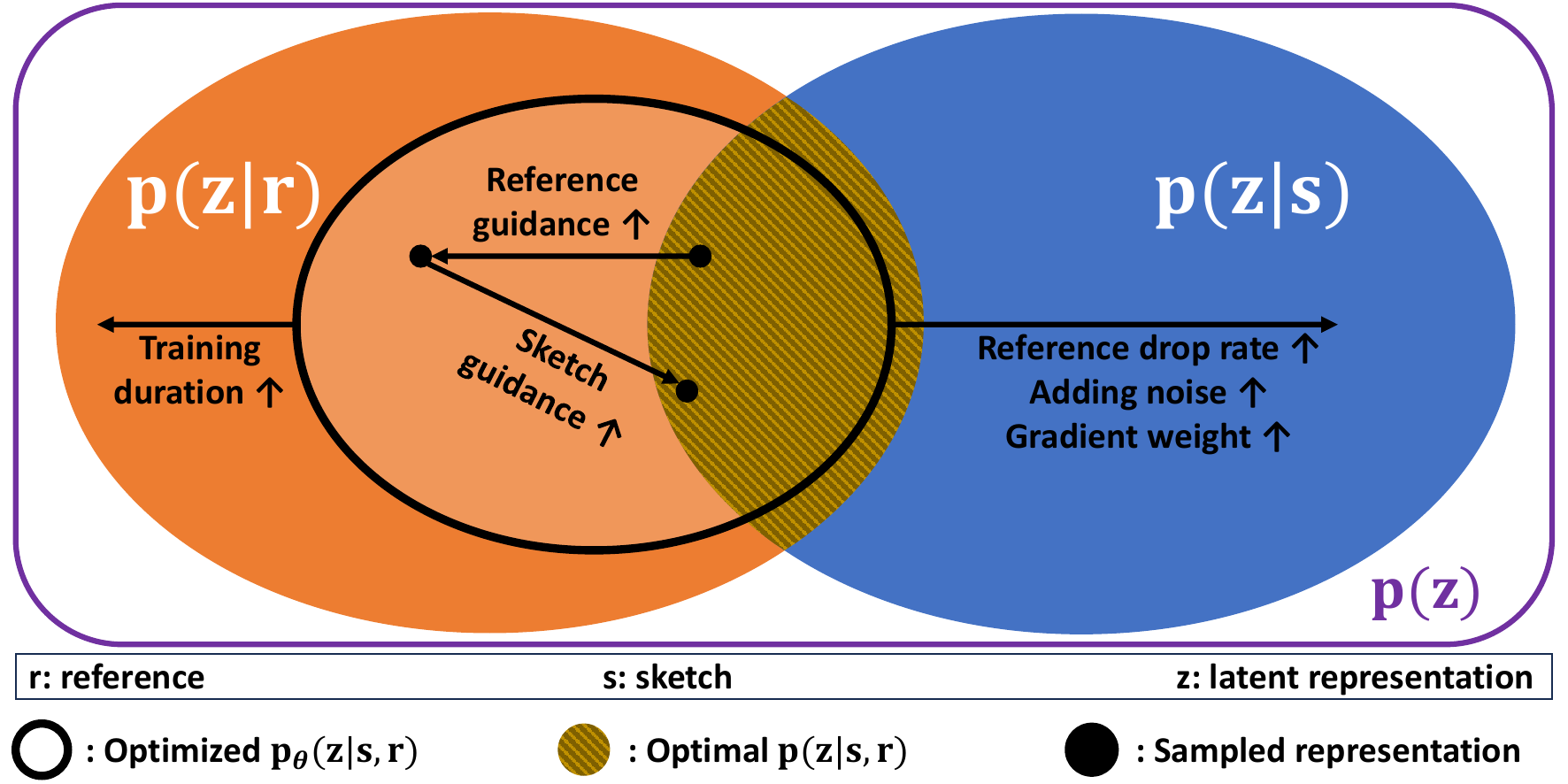}
	\caption{Illustration of the distribution problem. Most parts of the optimized distribution $p_{\theta}(z|s,r)$ after training lie outside of $p(z|s)$.}  
	\label{actdistributions}
\end{figure}

\begin{figure*}[ht]
	\centering
	\includegraphics[width=1\linewidth]{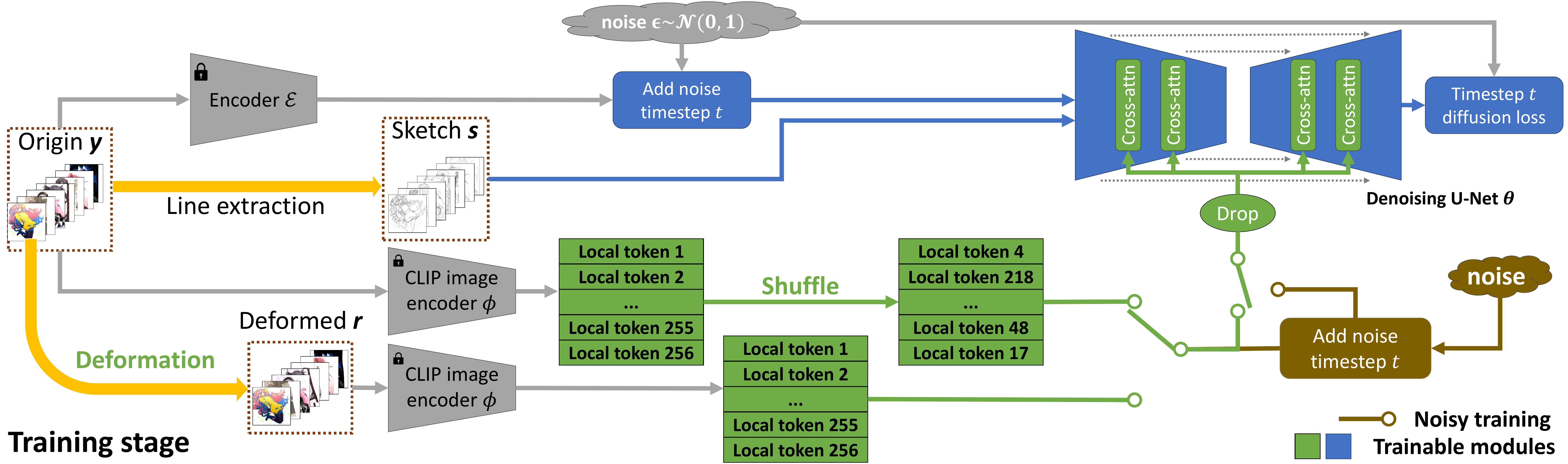}
	\caption{Training pipelines of the proposed \textit{Attention} models. We introduce two training strategies for the \textit{Attention} model, namely, deformation and shuffle training. Deformed images and sketch images are generated before training begins. Noisy training performs diffusion on the local tokens and is combined with either shuffle training or deformation training.}
	\label{training}
\end{figure*}
\begin{figure}[ht]
	\centering
	\includegraphics[width=1\linewidth]{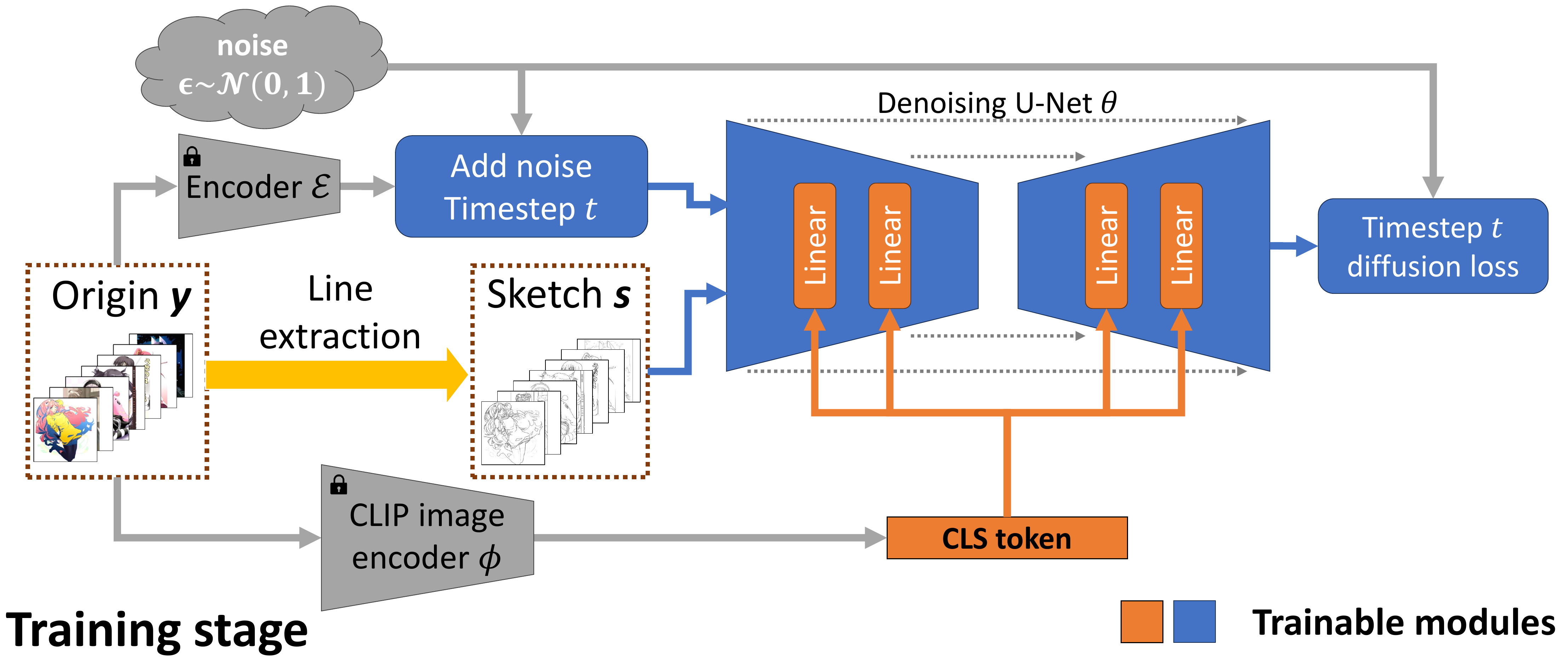}
	\caption{Training pipelines of the \textit{CLS} model.}
	\label{clstraining}
\end{figure}
3. The underlying reason stems from the distribution level, which is usually inevitable and also the major reason for the deterioration when training the whole network with both conditions jointly. When we train the dual-conditioned DM, there are two related conditional distributions, $p(z|s)$ and $p(z|r)$. We assume these distributions as ideal distributions, and images composed of features that are only inside the respective distributions are visually pleasant color images. Theoretically, if the generated images, which are sampled from the distribution $p_{\theta}(z|s,r)$, always remain within the distribution $p(z|s)$, their quality and segmentation should not be degraded by the newly introduced condition $r$; also, their semantic correspondence with the sketch should not be influenced. Nevertheless, we can observe notable deterioration by comparing rows (a),(b) with (c),(d),(e),(f) in Figure \ref{distribution-vis}, where results from two baseline methods show worse quality of textures and segmentation after introducing the reference conditions. This finding indicates that the actual distribution $p_{\theta}(z|s,r)$ of these models deviates from $p(z|s)$ and can be regarded as a kind of out-of-distribution (OOD). 

With our experimental results as a basis, we use Figure \ref{actdistributions} to illustrate the relationships among different distributions when training models with both conditions. When the optimized $p_{\theta}(z|s,r)$ is closer to $p(z|r)$, the segmentation of colorized images relies more on the reference images, and vice versa. Related experiments are discussed in Section 5. 

\subsection{Reference-based Training}
Our reference-based models are initialized using Waifu Diffusion \cite{waifudiffusion}, and a pre-trained CLIP Vision Transformer (ViT) from OpenCLIP-H \cite{RadfordKHRGASAM21,openclip,openclip-2,schuhmann2022laionb} is used to extract image tokens from reference images and remains frozen during training. For a $224\times224$ image, the CLIP ViT outputs 257 tokens, comprising 256 local tokens and one CLS token. The CLS token encapsulates the global semantic information of the reference image, while local tokens hold regional semantic content. We propose two reference-based models, \textit{CLS} and \textit{Attention}, differentiated by their token usage. Their training pipelines are illustrated in Figs. \ref{training} and \ref{clstraining}, respectively. The \textit{CLS} model leverages only the CLS token, replacing all cross-attention modules in the denoising U-Net with linear layers. The \textit{Attention} models utilize all local tokens for generation guidance, thereby maintaining an architecture similar to SD v1.5/2.1 \cite{RombachBLEO22}, the effectiveness of which in conditional generation has been demonstrated by various applications \cite{controlnet,ruiz2022dreambooth}.

\begin{figure*}[ht]
	\centering
	\includegraphics[width=1\linewidth]{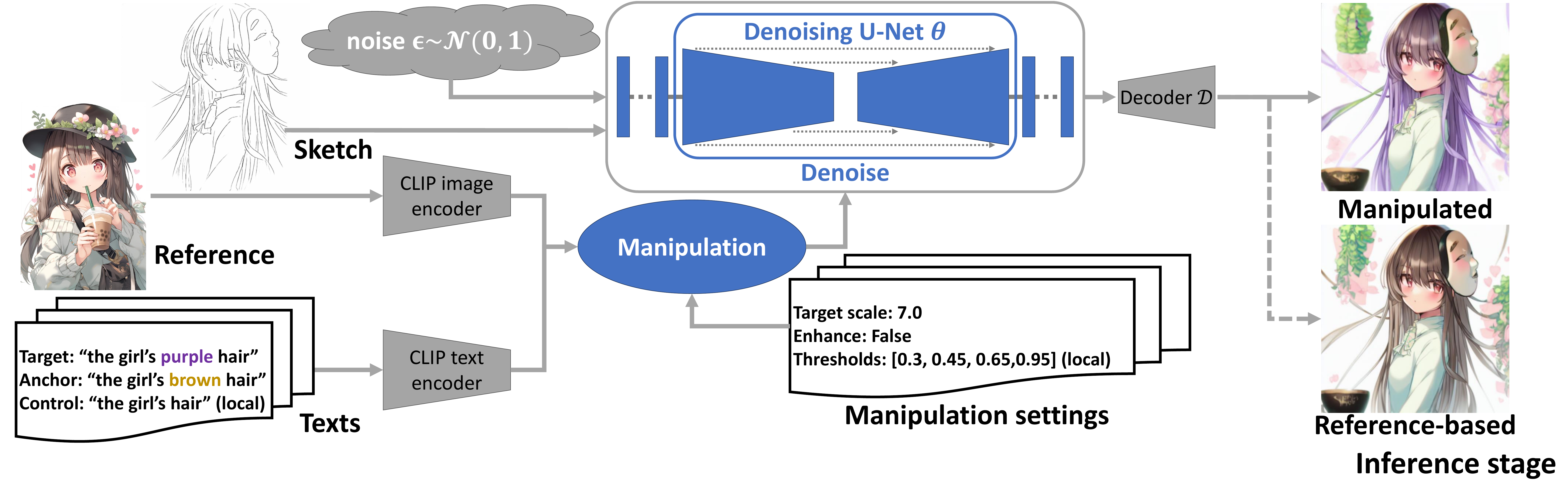}
	\caption{Our inference pipeline. The image tokens are edited before being input to the denoising U-Net. Illustrated results were generated by the \textit{Attention} model using local manipulation.}
	\label{inference}
\end{figure*}
Following \cite{controlnet}, we implement trainable convolutional layers in the denoising U-Net to downscale sketch inputs to the latent level, and these downscaled sketch features are added to the forward ones instead of being concatenated. The training of \textit{Attention} models requires additional processing for the reference inputs, so we accordingly adopt the following two processing schemes to obtain the reference inputs and train the \textit{Attention} model.

1. Deformation training: To address the data limitation, a widely adopted solution is to generate reference images from ground truth color images using deformation algorithms \cite{ZhangLW0L18,LeeKLKCC20,yan-cgf,abs-2303-11137}. In this paper, we utilize \cite{SchaeferMW06} to produce reference images before training. While this training method ameliorates the distribution issue from one perspective, it simultaneously degrades the quality of the generated images.

2. Latent shuffle training: Generating reference images can be time-consuming and storage-intensive. To avoid the possible impact caused by the spatial correspondence, we swap the sequence of local tokens before inputting them to the U-Net, as shown in Figure \ref{training} \cite{OordVK17,EsserRO21}.

Models trained by the respective scheme are labeled by \textit{Deform} and \textit{Shuffle} in the following sections. The diffusion loss for vanilla reference-based training is defined as
\begin{equation}
	\mathcal{L}(\theta)=\mathbb{E}_{\mathcal{E}(y),\epsilon,t,s,r}[\|\epsilon-\epsilon_{\theta}(z_{t},t,s,\tau_{\phi}(r))\|^{2}_{2}],
	\label{finetune}
\end{equation} 
where $\phi$ and $\tau_{\phi}$ denote the CLIP ViT and extracted tokens, respectively. Compared to deformation-trained counterparts, shuffle-trained models can generate results with a more vivid texture, although they are more likely to suffer from deterioration in segmentation due to the distribution problem. Therefore, most of our models were trained using latent shuffle to investigate the effectiveness of the proposed methods in mitigating the distribution problem.

\subsection{Solutions to the Distribution Problem}
To mitigate the distribution problem among \textit{Attention} models, we propose three solutions to move the optimized $p_{\theta}(z|s,r)$ towards $p(z|s)$, as explained in Section 3.2.

The first method, termed dropping training, randomly drops reference inputs during training with a drop rate much higher than 0.2, a suggested value in \cite{abs-2207-12598}. This slows down the optimization of cross-attention modules, thereby enabling the network to generate fine-grained textures before the optimized distribution $p_{\theta}(z|s,r)$ is out of $p(z|s)$. Default reference drop rates are empirically set to 0.75 for deformation training and 0.8 for shuffle training. 

The second method, called noisy training, is identified by the brown switch in Fig. \ref{training}. The noisy training tackles the distribution problem from all angles introduced in Section 3.2 by dynamically adding noise to local tokens in accordance with the timestep $t$. As reported by \cite{zhang2023prospect}, many low-level features, which are color-related, are determined in the early stages of denoising and can be disentangled from other embeddings. Therefore, reducing the semantics of the reference embedding, particularly in the early steps, facilitates the disentanglement of color-related embeddings. Meanwhile, as the reference embeddings are noised, the semantics they contain become much less pronounced and no longer align well with those of the ground truth. This avoids the deterioration of LDM and makes its distribution closer to $p(z|s)$. The objective function of noisy training is formulated as 
\begin{equation}
	\mathcal{L}(\theta)=\mathbb{E}_{\mathcal{E}(y),\epsilon,t,s,r}[\|\epsilon-\epsilon_{\theta}(z_{t},t,s,\tau_{\phi,t}(r))\|^{2}_{2}],
	\label{finetune}	
\end{equation}
where $\tau_{\phi,t}(r)=\alpha_{t}\tau_{\phi}(r)+\beta_{t}\epsilon_{r}$ and $\epsilon_{r}\sim\mathcal{N}(0,1)$. Compared to other solutions, this method significantly diminishes the distribution problem.

The main goal of the dropping training is to enable the network to generate $\epsilon_{t}$ satisfying $z_{t}\in p_{\theta}(z_{t}|z_{t+1},s,t)$. To better understand the distribution problem, we propose dual-conditioned training, which directly penalizes the difference between the sketch-based results and the ground truth. The dual-conditioned loss is accordingly organized as follows:
\begin{equation}
	\begin{aligned}
		\mathcal{L}(\theta)=\mathbb{E}_{\mathcal{E}(y),\epsilon,\epsilon',t,s,r}[&\|\epsilon-\epsilon_{\theta}(z_{t},t,s,\tau_{\phi}(r))\|^{2}_{2}+\\
		&\lambda\|\epsilon'-\epsilon_{\theta}'(z'_{t},t,s)\|^{2}_{2}],
	\end{aligned}
	\label{dual-loss}
\end{equation}
where $z_{t}$ and $z'_{t}$ are diffused from $z_{0}$ using different noises $\epsilon$ and $\epsilon'$, respectively, and $\lambda$ is set to 4 by default. In the following sections, models trained using the dropping, noisy, and dual-conditioned methods are referred to as the \textit{Drop} model, \textit{Noisy} model, and \textit{Dual} model, respectively.

Our experimental results (presented in Section 5) indicated that, far away from the ideal distribution $p(z|s)$, textures inside the optimized $p_{\theta}(z|s)$ were much coarser than those of $p_{\theta}(z|r)$. Therefore, in order to ensure the network is capable of generating fine-grained textures and suffers less from the deterioration caused by the distribution problem, we need to carefully decide the training duration, drop rate, and $\lambda$ used in Eq. \ref{dual-loss} for dropping training and dual-conditioned training. 

Overall, we consider noisy training as the most promising solution to the distribution problem, and we accordingly trained the \textit{Shuffle-noisy} model longer to investigate its effectiveness. However, it is important to note that the \textit{Noisy} model still suffers from the distribution problem caused by the semantic alignment of data.
        \section{Text-based Manipulation}
Compared to T2I models, adjusting the prompt conditions is more difficult for image-guided networks. We accordingly adopt a zero-shot interpolation method for the proposed \textit{CLS} model. DALL-E-2 \cite{abs-2204-06125} has demonstrated that an image-guided model utilizing CLIP encoders can modify outputs gradually using normalized text embedding. Therefore, we can also adjust image embeddings to align with the target degree of visual attributes specified by texts before inputting them to the denoising U-Net $\theta$. The inference pipeline is illustrated in Figure \ref{inference}.
\subsection{Global Text-Based Manipulation}
\begin{algorithm}[t]
	\caption{Sequential global manipulation.}
	\KwIn{CLS token: $\vec v_{cls}$\\
		\:\quad\qquad Normalized embeddings of target prompts: $\vec e[1..N]$ \\
		\:\quad\qquad Normalized embeddings of anchor prompts: $\vec a[1..N]$ \\
		\:\quad\qquad Target scales: $target\_scale[1..N]$ \\
		\:\quad\qquad Enhance flags: $enhance[1..N]$}
	\For {$i=1,2,..,N$}
	{
		\If{$\vec a[i]\,is\,not\,null$} {
			\If{$enhance[i]\,is\,true$} {
				$\vec{v}_{cls}\leftarrow\vec{v}_{cls}-(\vec v_{cls}\cdot\vec a[i])*\vec a[i]$\\
				$\vec{v}_{cls}\leftarrow\vec{v}_{cls}+(target\_scale[i]-\vec v_{cls}\cdot\vec e[i])*\vec e[i]$
			}
			\Else{
				$\vec v_{cls}\leftarrow\vec v_{cls}+target\_scale[i]*(\vec e[i]-\vec a[i])$
			}
		}
		\Else{
			\If{$enhance[i]\,is\,true$} {
				$\vec v_{cls}\leftarrow\vec v_{cls}+target\_scale[i]*\vec e[i]$
			}
			\Else{
				$\vec v_{cls}\leftarrow\vec v_{cls}+(target\_scale[i]-\vec v_{cls}\cdot\vec e[i])*\vec e[i]$
			}
		}
	}
	\Return $\vec v_{cls}$
	\label{global manipulation}
\end{algorithm}
The CLIP score is widely used to evaluate the correlation between a generated image and a given caption. It is calculated as the projection of the image CLS token onto the text CLS token. While using image tokens as prompt inputs, we can directly modify the generated results using this projection-based correlation. To simplify the expression, we denote the extracted image tokens (previously represented as $\tau_{\phi}(r)$) and the normalized text CLS token as vectors $\bm{\vec v}$ and $\vec e$, respectively. Specifically, the CLS token is denoted as $\vec v_{cls}$, and we can calculate the modified CLS token $\vec v^{m}_{cls}$ as
\begin{equation}
	\vec v^{m}_{cls}=
	\begin{cases}
		\vec v_{cls}+target\_scale*\vec e & \mbox{$enhance$}\\		
		\vec v_{cls}+(target\_scale-\vec v_{cls}\cdot\vec e)*\vec e & \mbox{$not~enhance$}\\
	\end{cases},
	\label{adjust}
\end{equation} 
where $target\_scale$ and $enhance$ are user-defined parameters. They indicate the target scale of the interpolation and whether the manipulation should be enhanced to achieve a more obvious change, respectively. Similar to DALL-E-2, the manipulation can be improved through the normalized embedding of an anchor text, termed $\vec a$. The first method, where $enhance$ is set to false, calculates $\vec v^{m}_{cls}$ with the anchor text as
\begin{equation}
	\vec v^{m}_{cls}=\vec v_{cls}+target\_scale*(\vec e-\vec a).
\end{equation}
The global manipulation can be further enhanced by first eliminating the anchor attribute with $\vec a$ before adding $\vec e$. The modified CLS token $\vec v'_{cls}$ is then calculated as
\begin{equation}
	\begin{aligned}
	&\vec v'^{m}_{cls}=\vec v_{cls}-(\vec v_{cls}\cdot\vec a)*\vec a,\\
	&\vec v^{m}_{cls}=\vec v'^{m}_{cls}+(target\_scale-\vec v'^{m}_{cls}\cdot\vec e)*\vec e.
	\end{aligned}
\end{equation}
However, enhancing the manipulation with an anchor text would make unrelated attributes more likely to be jointly changed. The sequential manipulation of $\vec v_{cls}$ is shown in Algorithm \ref{global manipulation}. The target scales ranging proposed in $[4,15]$ can generate reasonable results.

\subsection{Local Text-Based Manipulation}
As \textit{Attention} models utilize local tokens as conditions, global manipulation becomes ineffective due to the absence of spatial information. Accordingly, we propose a semi-automatic algorithm for local tokens to accomplish manipulation. Note that, to ensure the capability of accepting arbitrary text as input, the proposed local manipulation remains zero-shot.
\begin{figure}[t]
	\centering
	\includegraphics[width=1\linewidth]{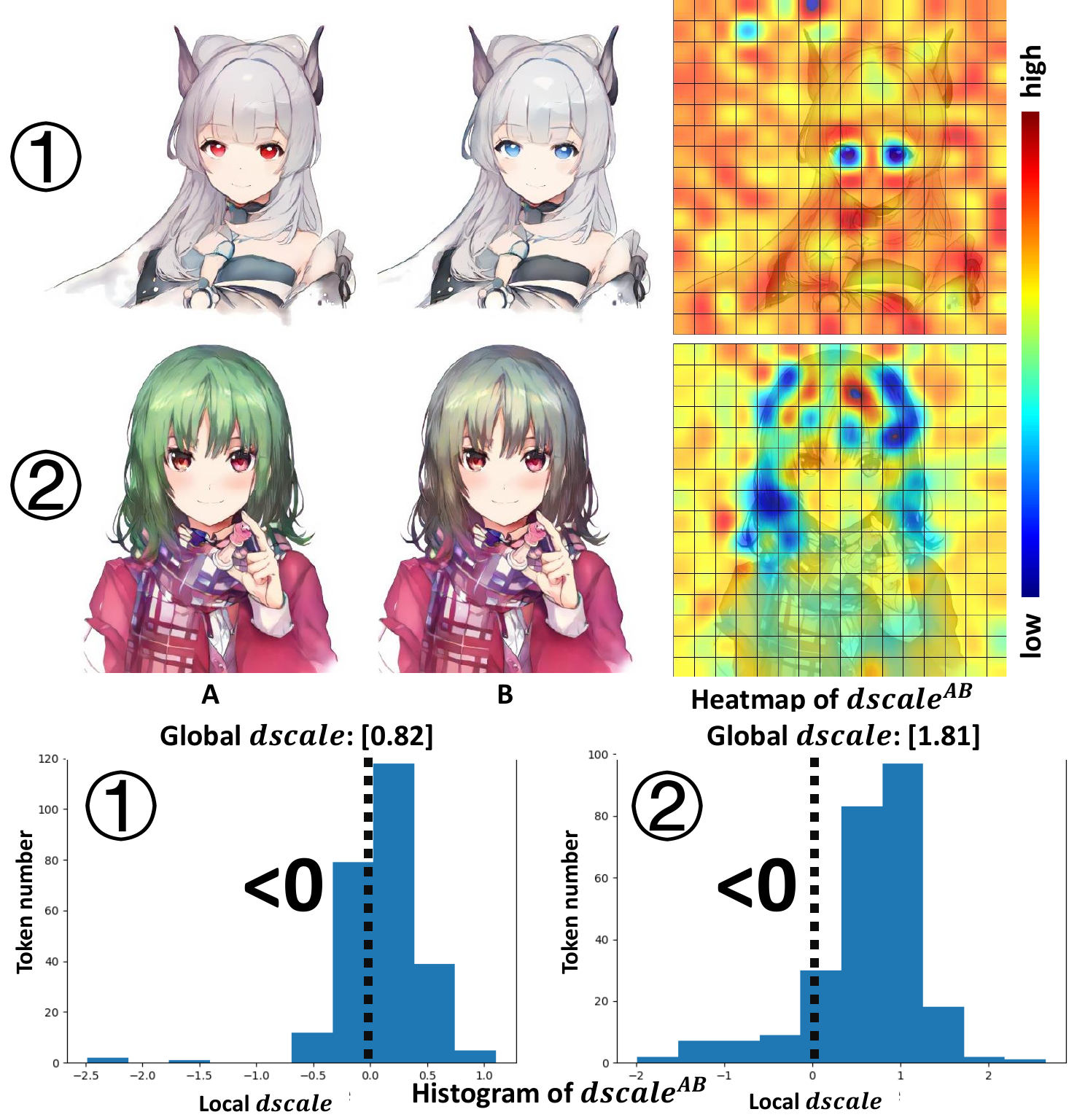}
	\caption{Visualization of $\bm{dscale}^{AB}$ corresponding to the texts ``the girl's red eyes'' (upper) and ``the girl's green hair'' (lower), respectively.}
	\label{projections}
\end{figure}

We first introduce three terms used in the proposed local manipulation: $dscale$, Position Weight Vector (PWV) $\bm{m}$, and PWV $\bm{\omega}$. We already know that the correlation between an image and a caption can be evaluated through the CLIP projection, formulated as $corr=\vec v_{cls}\cdot\vec e$. We have observed that the local tokens also demonstrate the ability of zero-shot segmentation, which suggests that such correlation is also computable using local tokens. Therefore, we extend the calculation of the correlation vector as $corr_{i}=\vec v_{i}\cdot\vec e$, with $i\in\{cls,1,2,..,n\}$ and $n$ being the total number of local tokens, which is 256 for the adopted OpenCLIP-H, and define $dscale_{i}^{AB}=\vec v_{i}^{A}\cdot\vec e-\vec v_{i}^{B}\cdot\vec e$. Our aim is to use $dscale_{cls}$ and PMVs $\bm{m},\bm{\omega}$ to simulate $\bm{dscale}^{AB}$, where $\bm{dscale}^{AB}=[dscale_{1}^{AB},..,dscale_{n}^{AB}]$. If the difference between images A and B can be fully described using the text embedding $\vec e$, we can approximate $\bm{\vec v}^{A}$ as 
\begin{equation}
    \bm{\vec v}^{A}=\bm{\vec v}^{B} + \bm{dscale}^{AB}
    \label{approx}
\end{equation}
In our observations, we noticed that the local and CLS tokens exhibit different directional changes when projected onto the text embedding. We find that for the given text ``a girl with green hair'', as the hair becomes greener, the projection of the CLS token along the text embedding direction lengthens, which is labeled as $corr$ on top of the histograms in Figure \ref{projections}. Conversely, the projections of the most relevant local tokens decrease, while those of irrelevant tokens increase. These dynamics can be observed from the heatmaps of $\bm{dscale}^{AB}$, where regions closely related to the text are marked in blue. Given that blue is used to represent lower values, the heatmaps clearly indicate that the $\bm{dscale}^{AB}$ values for these regions are negative, as corroborated by the histograms.

\begin{figure}[t]
	\centering
	\includegraphics[width=1\linewidth]{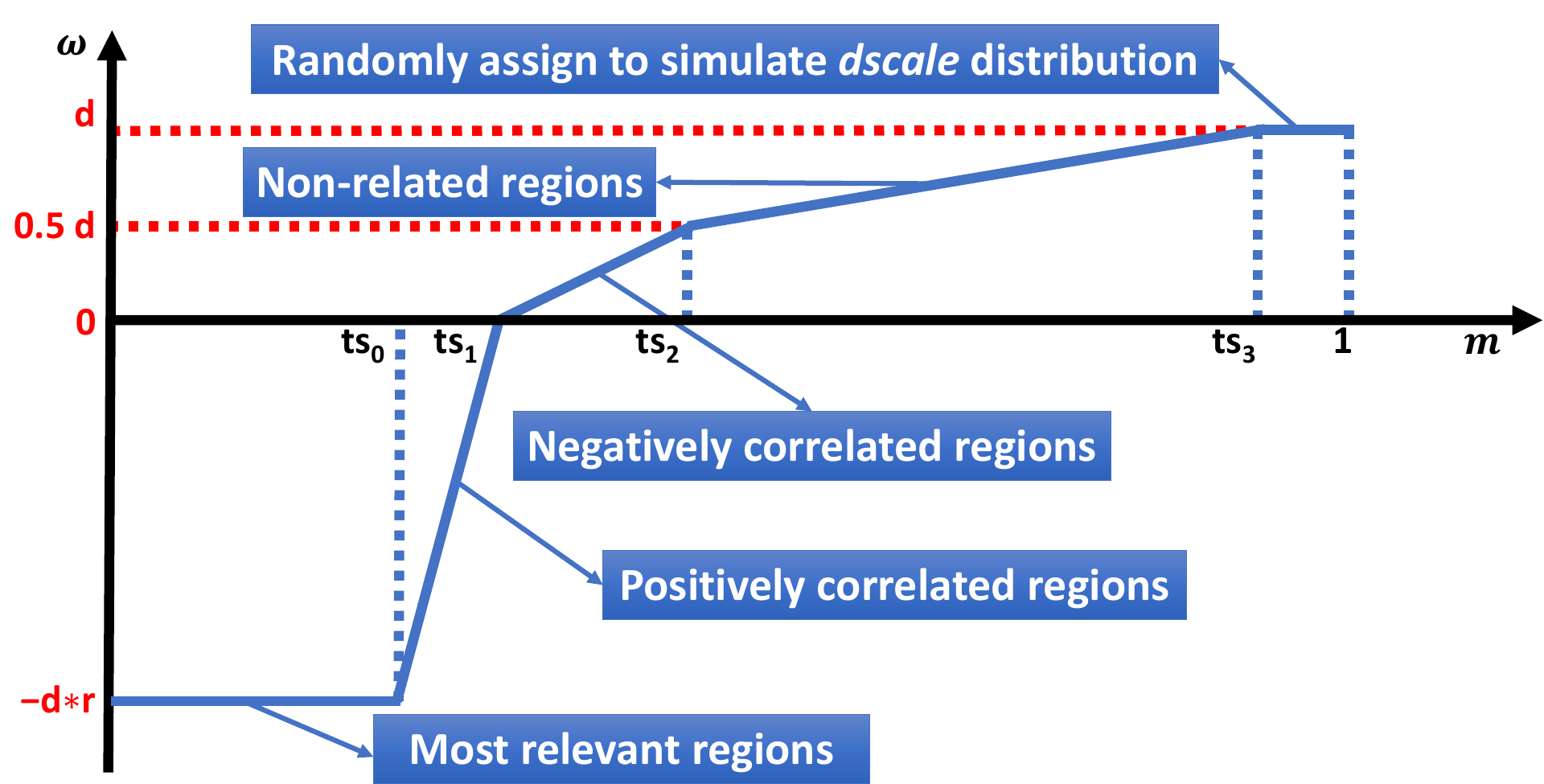}
	\caption{Plotting $\omega_{i}$ as a function of $m_{i}$ in Eq. \ref{calculate_m}. We divide the domain into five intervals to reduce the influence of the manipulation on unrelated attributes.}
	\label{curve}
\end{figure}
We use the control prompt whose embedding is denoted as $\vec c$ to locate the region of local manipulation and calculate the PWV $\bm{m}$ as
\begin{equation}
    \bm{m}=\mathcal{F}(\bm{\vec v}\cdot\vec c),
    \label{definition_m}
\end{equation}
where $\mathcal{F}$ indicates the min-max normalization. By leveraging the correlation PWV $\bm{m}$, we formulate the PWV $\bm{\omega}$ as
\begin{equation}
	\omega_{i}=
	\begin{cases}
		-d*r, &\mbox{$m_{i}\leqslant ts_{0}$}\\
		-d*r+d*r*\frac{m_{i}-ts_{0}}{ts_{1}-ts_{0}}., &\mbox{$ts_{0}<m_{i}\leqslant ts_{1}$}\\
		0.5*d*\frac{m_{i}-ts_{1}}{ts_{2}-ts_{1}}, &\mbox{$ts_{1}<m_{i}\leqslant ts_{2}$}\\
		0.5*d+0.5*d*\frac{m_{i}-ts_{2}}{ts_{3}-ts_{2}}, &\mbox{$ts_{2}<m_{i}\leqslant ts_{3}$}\\
		d, &\mbox{$m_{i}>ts_{3}$}\\
	\end{cases}
	\label{calculate_m}
\end{equation}
where $m_{i}$ and $\omega_{i}$ represent the i-th element of $\bm{m}$ and $\bm{\omega}$, respectively, with $i\in\{1,..,n\}$. We illustrate this function in Figure \ref{curve}. In this equation, $d$ is computed as
\begin{equation}
	d=
	\begin{cases}
		target\_scale-\vec v_{cls}\cdot\vec a, &\mbox{$enhance$}\\
		target\_scale-\vec v_{cls}\cdot\vec e. &\mbox{$not~enhance$}\\
	\end{cases}.
	\label{calculate_d}
\end{equation}
The hyperparameters $r$ and $ts_{i}$ in Eq. \ref{calculate_m} denote the strength ratio for the most pertinent areas and the thresholds for differentiating all areas of the image, respectively. The rough definitions of different threshold intervals are given in Figure \ref{curve}. The default settings for the hyperparameters $r$ and $[ts_{0},ts_{1},ts_{2},ts_{3}]$ are $2$ and $[0.5, 0.55, 0.65, 0.95]$, respectively. We set four thresholds to reduce the manipulation's influence on irrelevant visual attributes as much as possible. Experimentally, target visual attributes should be encompassed within the regions defined by $\bm{m}\leqslant ts_{1}$, while attributes intended for preservation should be within the $\bm{m}>ts_{2}$ region. Accordingly, we can formulate the adjustment equation for the local tokens as
\begin{equation}
	\bm{\vec v}^{m}=\bm{\vec v}+(\bm{\omega}+\bm{\beta}*\bm{\vec v}\cdot\vec a)*(\vec e-\vec a),
\end{equation}
where $\bm{\beta}$ corresponds to the $enhance$ flag. If there is no anchor prompt, the equation is reorganized as
\begin{equation}
	\bm{\vec v}^{m}=\bm{\vec v}+\bm{\omega}*\vec e.
\end{equation}
This formulation is similar to Eq. \ref{approx}. This calculation can also be expanded to enable the sequential manipulation of multiple text pairs, as detailed in Algorithm \ref{local manipulation}. Nevertheless, defining suitable thresholds for a control prompt can be challenging. To alleviate this difficulty, we have designed an interactive user interface that visually assists users in identifying the regions selected by each threshold. Implementation of the proposed manipulation is included in the supplementary materials.

\begin{algorithm}[t]
\SetAlgoNoLine
	\KwIn{
		Local tokens: $\bm{\vec v}$; CLS token: $\vec v_{cls}$\\
		\:\quad\qquad Normalized embeddings of target prompts: $\vec e[1..N]$\\
		\:\quad\qquad Normalized embeddings of anchor prompts: $\vec a[1..N]$\\
		\:\quad\qquad Normalized embeddings of control prompts: $\vec c[1..N]$\\
		\:\quad\qquad Target scales: $target\_scale[1..N]$\\
		\:\quad\qquad Enhance flags: $enhance[1..N]$ \\
		\:\quad\qquad Thresholds list: $ts_{0,..,3}[1..N]$\\
		\:\quad\qquad Strength factor: $r$}
	\For {$i=1,2,..,N$}{
		\If{$\vec a[i]\,is\,not\,null$} {
			\If{$enhance[i]\,is\,true$} {
				$d\leftarrow target\_scale[i]-\vec v_{cls}\cdot\vec a[i]$\\
				$\bm{\beta}\leftarrow\bm{1}$
			}
			\Else{
				$d\leftarrow target\_scale[i]-\vec v_{cls}\cdot\vec e[i]$\\
				$\bm{\beta}\leftarrow\bm{0}$
			}
			$\bm{m}\leftarrow\mathcal{F}(\bm{\vec v}\cdot\vec c[i])$\\
			$\bm{\omega}\leftarrow\bm{\omega}(\bm{m},d,ts_{0,..3}[i], r)$ according to Eq \ref{calculate_m}\\
			$\bm{\vec v}\leftarrow\bm{\vec v}+(\bm{\omega}+\bm{\beta}*\bm{\vec v}\cdot\vec a)*(\vec e[i]-\vec a[i])$
		}
		\Else{
			$d\leftarrow target\_scale[i]$\\
			$\bm{m}\leftarrow\mathcal{F}(\bm{\vec v}\cdot\vec c[i])$\\
			$\bm{\omega}\leftarrow\bm{\omega}(\bm{m},d,ts_{0,..3}[i], r)$ according to Eq \ref{calculate_m}\\
			$\bm{\vec v}\leftarrow\bm{\vec v}+\bm{\omega}*\vec e[i]$
		}
	}
	\Return $\bm{\vec v}$
 \caption{Sequential local manipulation.}
 \label{local manipulation}
\end{algorithm}
	\setcounter{figure}{10}
\begin{figure*}[t]
	\centering
	\includegraphics[width=1\linewidth]{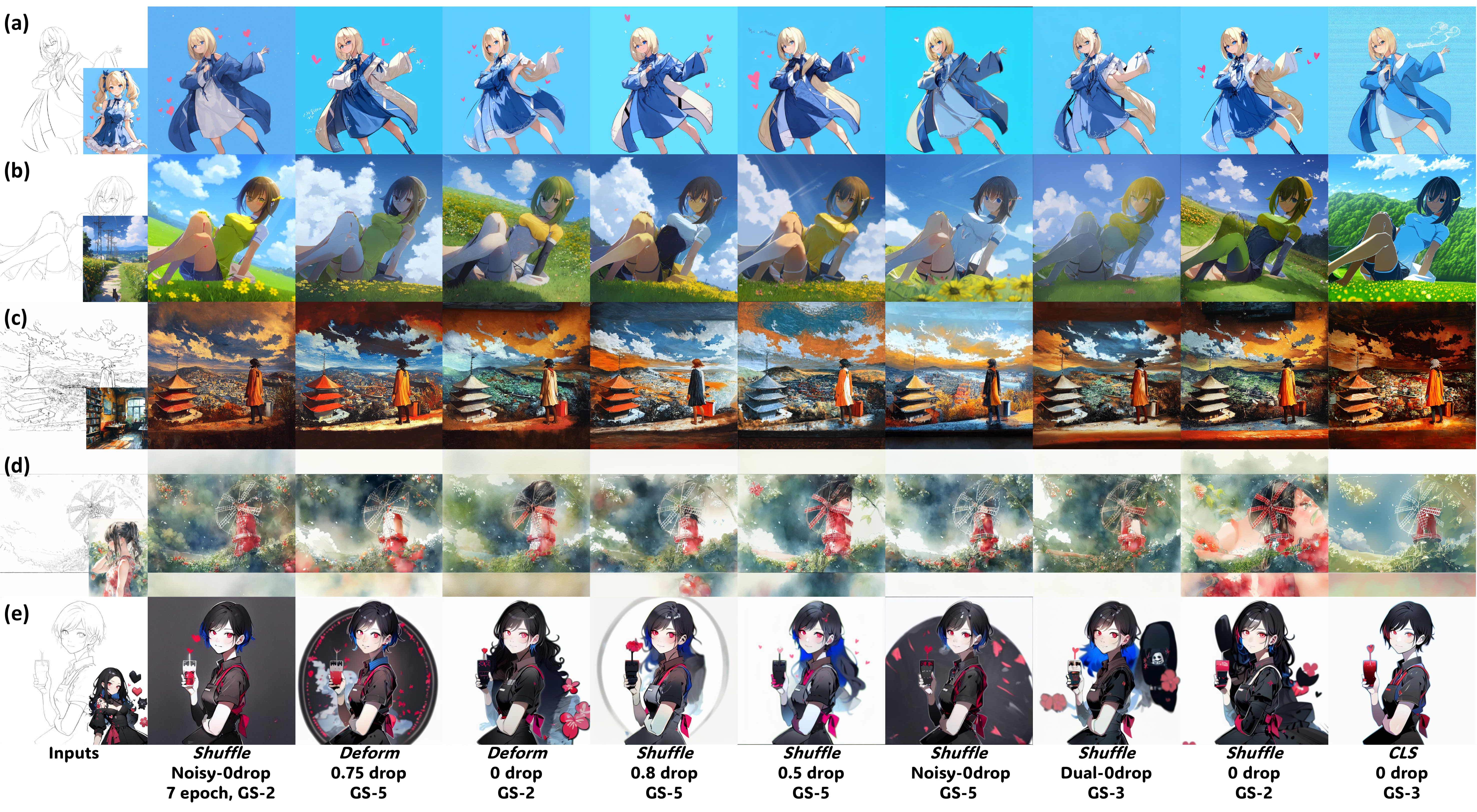}
	\caption{Colorized results generated by ablation models. As demonstrated here, the \textit{Shuffle-noisy} model is able to maintain semantic fidelity to the sketch input, even after extended training. Therefore, it is selected as our default model in subsequent comparisons with baseline methods.}
	\label{ablation-qualitative}
\end{figure*}
\setcounter{figure}{9}
\begin{figure}[t]
	\centering
	\includegraphics[width=1\linewidth]{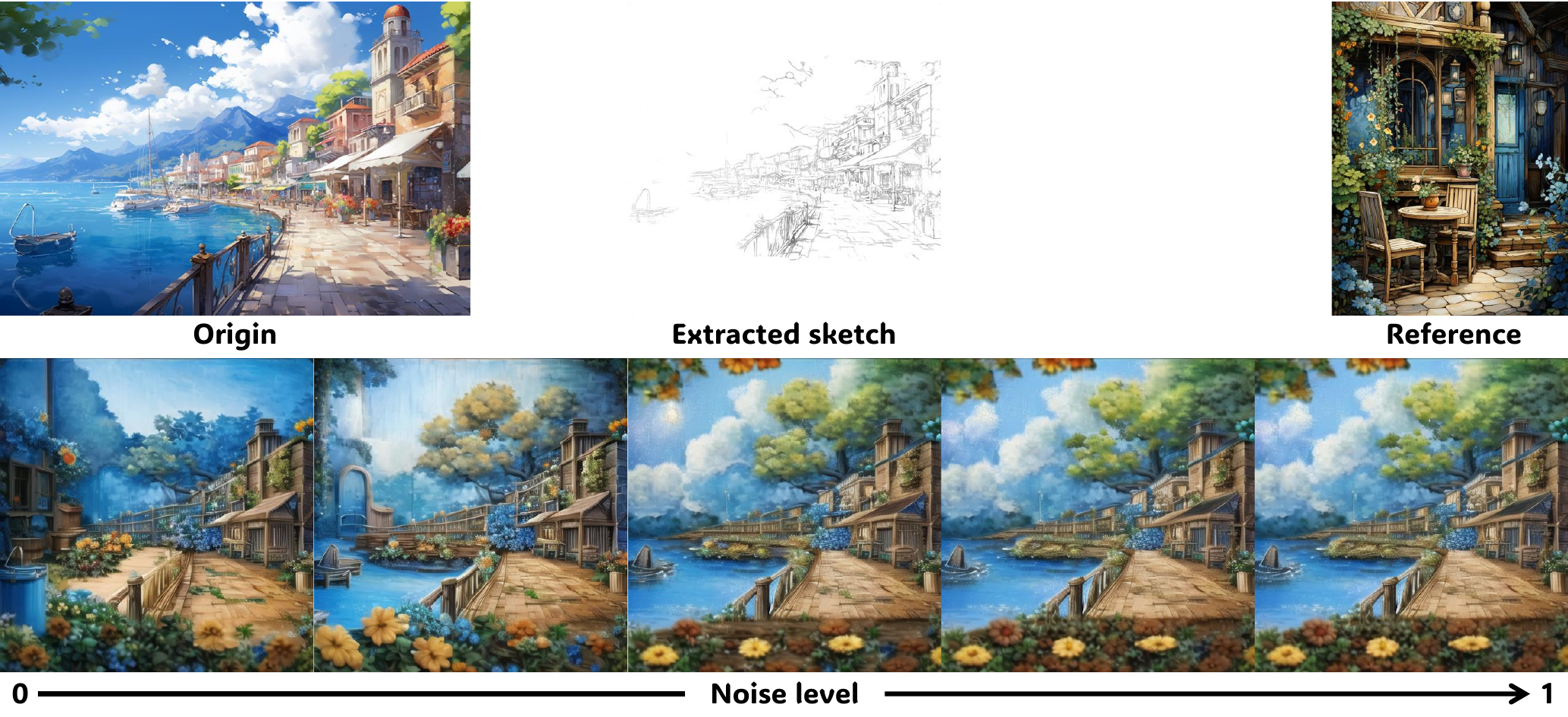}
	\caption{Illustration of the noisy sampling, which can increase the semantic fidelity to the sketch input without significantly degrading the quality of generated textures when combined with the noisy training.}
	\label{noisy-sampling}
\end{figure}
\setcounter{figure}{11}
\section{Experiment}
In this section, we first introduce a special sampling method in Section 5.1 and detail our implementation in Section 5.2. We then experimentally compare the proposed models through ablation studies in Section 5.3 and compare them to baselines in Section 5.4. We present our text-based manipulation in Section 5.5, followed by the results of a corresponding user study in Section 5.6. The Fréchet Inception Distance (FID) \cite{HeuselRUNH17,Seitzer2020FID} estimates the distribution distance between generated images and real images and is thus utilized to evaluate the performance of generative models in this section. However, as per our experiments, FID cannot subjectively reflect the distribution problem; therefore, qualitative results are considered more significant for our evaluation.

\subsection{Implementation Details}
\noindent\textbf{Noisy Sampling.} We introduce a special sampling method called ``noisy sampling'', which is achieved by adding noise to the local tokens according to the timestep $t$ and a hyperparameter $noise\_level$. In the proposed noisy sampling, the reference embeddings utilized in each denoising step $t$ are calculated as
\begin{equation}
    \tau_{\phi,t}(r)=
    \begin{cases}
        \alpha_{t}\tau_{\phi}(r)+\beta_{t}\epsilon_{r} & \mbox{if $(1-\frac{t}{T+0.0001})< noise\_level$}\\
        \tau_{\phi}(r) & \mbox{else}\\
    \end{cases},
\end{equation}
where $T$ is the total number of sampling steps and $noise\_level\in[0,1]$. Noisy sampling reduces the influence of reference embeddings in low-level features and correspondingly increases the semantic fidelity to the sketch input. An example is given in Figure \ref{noisy-sampling}. Note that, to better evaluate the distribution problem, noisy sampling was not used for all the comparisons illustrated in this paper.\\

\noindent\textbf{Training and Testing.} We implemented our models using PyTorch and trained them on an NVIDIA DGX-Station A100 with 4x NVIDIA A100-SXM 40G. The \textit{CLS} model and the \textit{Attention} models were trained for seven and five epochs on the training set, respectively, except for the \textit{Shuffle-noisy} model, which was also trained for seven epochs because the noisy training effectively disentangles spatial embeddings. The training of the \textit{Shuffle-Dual} model took eight days, whereas the training of the other models took approximately five days using Distributed Data-Parallel Training (DDP) and the AdamW optimizer \cite{KingmaB14,LoshchilovH19}. The training settings were as follows: learning\_rate = 1e-5, batch\_size\_per\_gpu = 10, betas = (0.9, 0.999), accumulative\_batches = 2, weight\_decay = 0.1. We adopted Stability-AI's official implementation of the DPM++ solver, which is multi-step and second-order \cite{0011ZB0L022,abs-2211-01095}, and our default number of sampling steps for testing was set to 20.\\

\noindent\textbf{Dataset.} We used Danbooru 2021 \cite{danbooru2021} as our original dataset to produce corresponding sketch and reference images. The sketch images were generated by jointly using SketchKeras \cite{sketchKeras} and Anime2Sketch \cite{xiang2022adversarial}, where the total training set includes 4M+ triples of (sketch, reference, color) images at a resolution of $512^{2}$. All quantitative evaluations were taken on a subset of Danbooru 2021, including 40,000+ ground truth tags and (sketch, color) image pairs. Samples of the training data are included in the supplementary materials.\\

\noindent\textbf{Dual Classifier-Free Guidance.} Our models can concurrently apply two forms of Classifier-Free Guidance (CFG) during inference, both of which set zero as the negative input. The guidance scales for reference-based and sketch-based guidance are denoted as GS and SGS, respectively, in subsequent sections.\\
\begin{table}[t]
	\centering
	\caption{FID scores for ablation models using variance preserving (VP) scheduler \cite{0011SKKEP21}. Drop rates are denoted by \{0, 0.5, 0.75, 0.8\}, indicating the specific rate used in training each model. Guidance scales for each validation are represented by \{GS-1, GS-2, GS-3, GS-5, GS-10\}. The top-performing score is emphasized in bold. \dag: Evaluated after seven epochs.}	
	\begin{tabular}{|c|c|c|c|c|c|}
		\hline
		\multicolumn{6}{|c|}{Fréchet inception distance (FID) $\downarrow$}  \\
        \hline
        \multicolumn{6}{|c|}{Ablation model}  \\
		\hline
		Model & GS-1 & GS-2 & GS-3 & GS-5 & GS-10\\
		\hline
		\textit{Deform-0} & 15.8590 & 10.8875 & 13.9459 & 20.7550 & 36.4256 \\
		\hline
		\textit{Deform-0.75} & 17.4646 & 12.9854 & 11.5916 & 11.7067 & 15.5636 \\
		\hline
		\textit{Shuffle-0} & 15.6971 & 10.3265 & 13.8398 & 22.1181 & 41.4941 \\
		\hline
        \textit{Shuffle-0.5} & 16.2813 & 10.7023 & 9.5553 & 9.4883 & 12.4227 \\
		\hline
		\textit{Shuffle-0.8} & 15.2748 & 10.5986 & 9.1956 & 9.2383 & 12.0642 \\
		\hline
		\textit{Noisy-0} & 15.5723 & 10.4629 & 9.0724 & \textbf{8.9314} & 11.5719 \\
		\hline
        \dag\textit{Noisy-0} & 11.7979 & 10.6517  & 12.2341 & 13.7150 & 16.5957 \\
        \hline
        \textit{Dual-0} & 18.8059 & 13.6929 & 13.2995 & 14.7224 & 25.2262 \\
        \hline
        \textit{CLS-0} & 13.5240 & 15.4600 & 19.9103 & 26.2609 & 41.8732 \\
        \hline
	\end{tabular}
	\label{fid}
\end{table}

Increasing the resolution for inference and applying Adaptive Instance Normalization (AdaIN) \cite{HuangB17} as well as attention injection \cite{lvmin-injection,abs-2305-18729,Tumanyan_2023_CVPR} can improve the similarity with references. Details can be found in the supplementary materials.

\subsection{Ablation Study}
As most baselines are not jointly trained with both conditions, and the semantic alignment of training data becomes the major factor contributing to the deterioration in both quality and segmentation, as stated in Section 3.2, comparison with ablation models is the most important part of our experiments. Since this deterioration cannot be adequately evaluated utilizing metrics, we conducted various qualitative comparisons to better observe this deterioration.\\
\begin{figure}[t]
    \centering
    \includegraphics[width=1\linewidth]{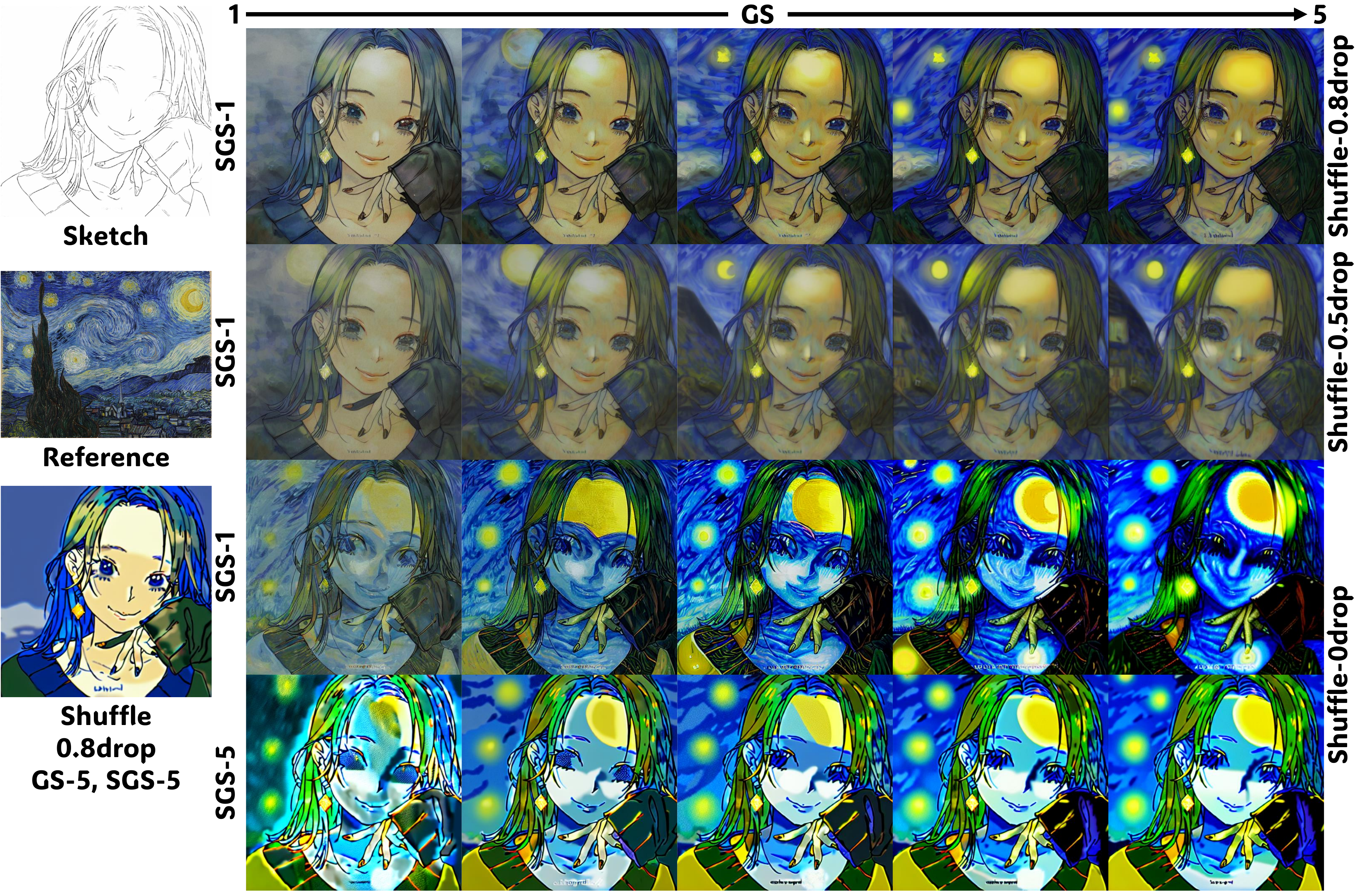}
    \caption{Results from ablation models trained using different drop rates. An increase in SGS makes the sampled features more likely to fall within $p_{\theta}(z|s)$, yielding visually more accurate segmentation but at the expense of fine-grained texture detail.}
    \label{cfg}
\end{figure}

\noindent\textbf{Training Strategy and Architecture.} We first evaluate the two variation models introduced in Section 3.3. As shown in Table \ref{fid}, \textit{Attention} models trained with different strategies achieved equivalent qualitative and quantitative results, demonstrating a better ability to transfer features than the \textit{CLS}. We can also observe from row (e) of Figure \ref{ablation-qualitative} that many ablation models erroneously rendered long hair. The results of the CLS model also demonstrate that the major deterioration of segmentation in Attention models is caused by the entangled spatial embeddings.
\begin{figure}[t]
	\centering
	\includegraphics[width=1\linewidth]{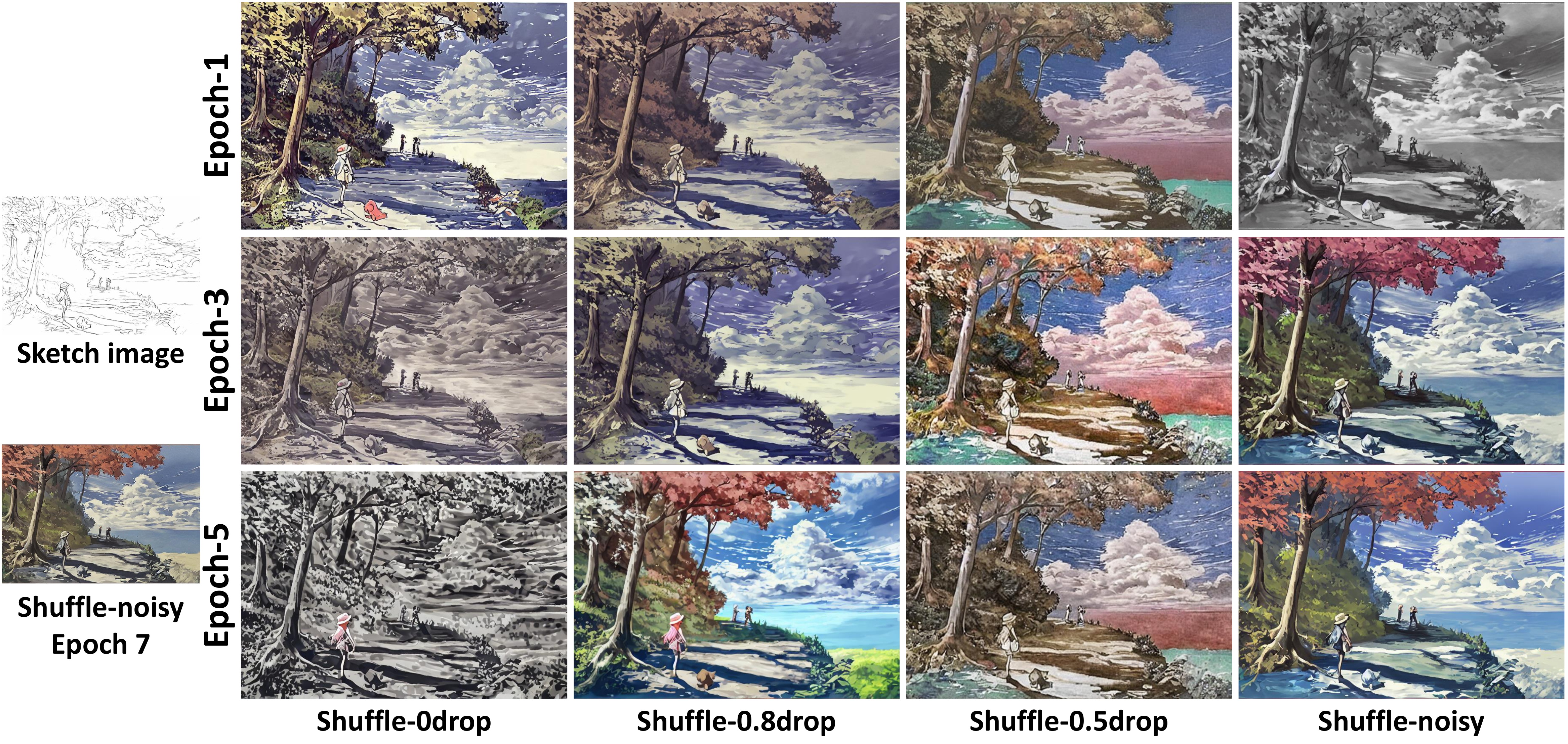}
	\caption{Colorization results without reference inputs, where the \textit{0drop} model fails to synthesize color very soon as the training progresses. SGS was set to 1.3 in this test.}
	\label{sketch}
\end{figure}
\begin{figure*}[t]
	\centering
	\includegraphics[width=1\linewidth]{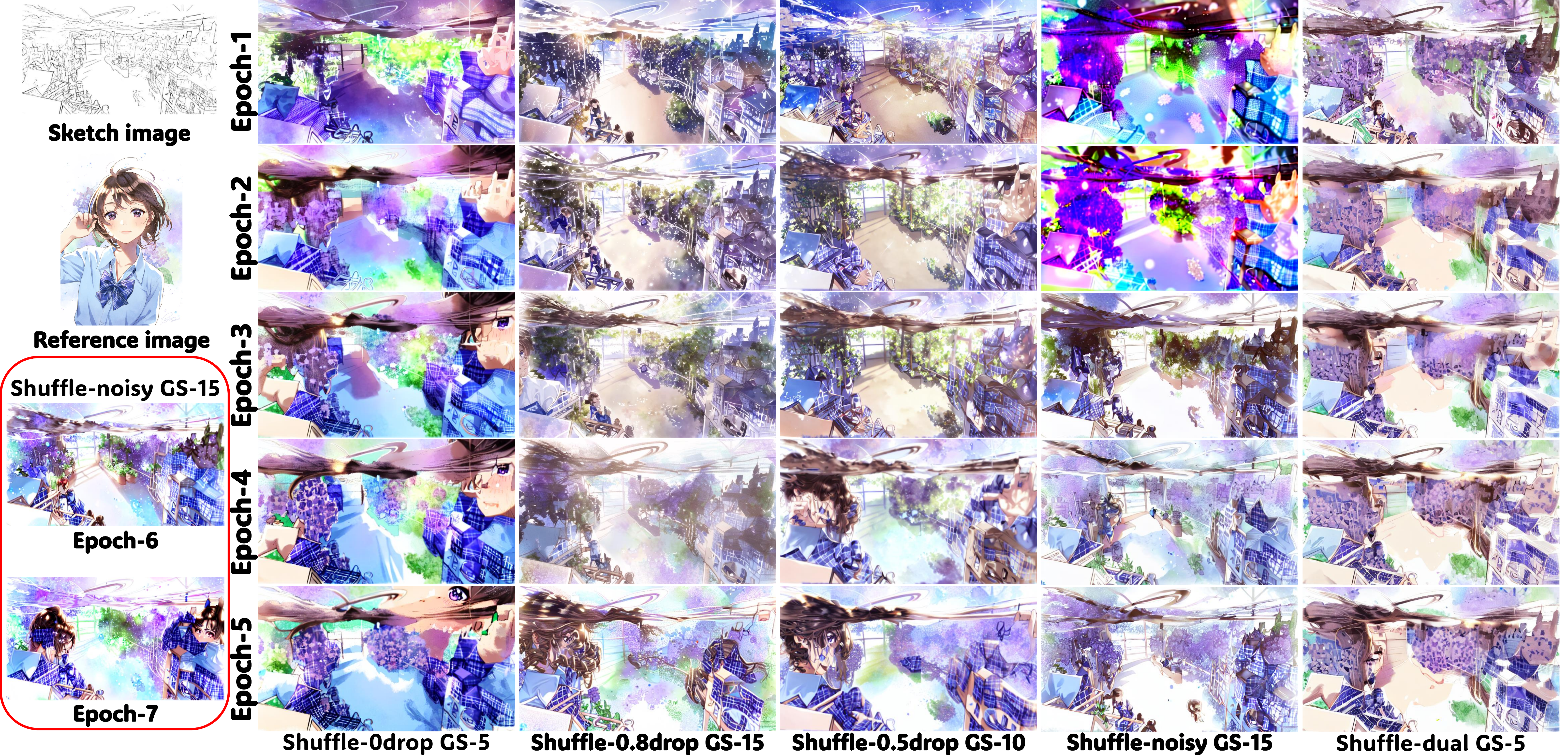}
	\caption{To better observe the distribution problem, we utilized the VP noise scheduler and extremely high reference guidance scales in this test. Aside from the \textit{Shuffle-noisy} model, all models generated significantly incompatible textures at Epoch 5.}
	\label{epoch}
\end{figure*}
\begin{figure*}[t]
	\centering
	\includegraphics[width=1\linewidth]{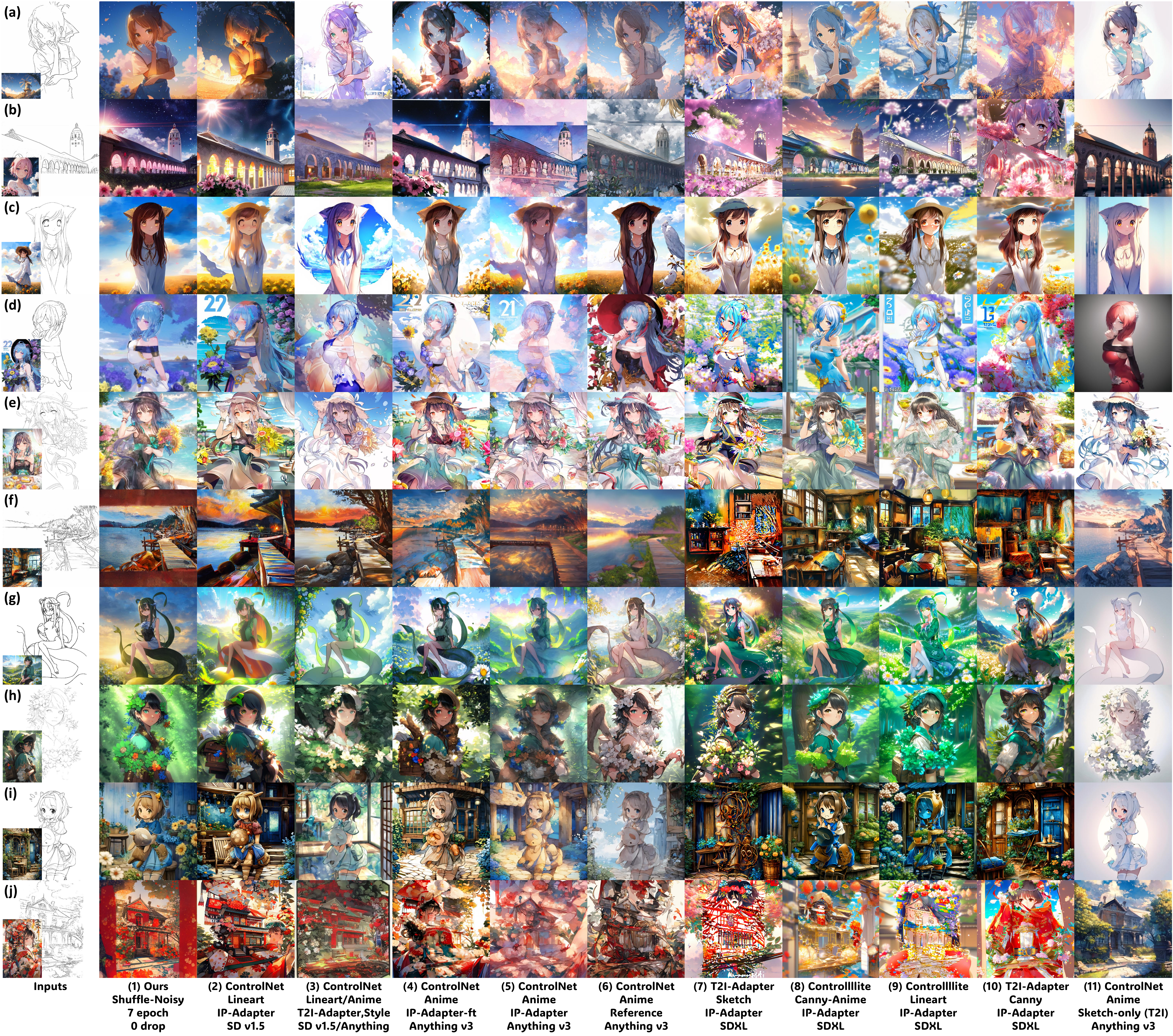}
	\caption{Qualitative comparison with baseline methods. We only adjusted GS for our method in this test, while most baseline methods necessitate precise adjustments of hyperparameters to obtain reasonable results without the distribution problem. Rows (h)--(j) display results where only the CFG scales were altered in baseline methods. Additionally, we fine-tuned \textit{IP-Adapter v1.5} with \textit{Anything v3} to align their distributions, labeled as \textit{IP-Adapter-ft}.}
	\label{baseline}
\end{figure*}
\begin{figure}[t]
    \centering
    \includegraphics[width=1\linewidth]{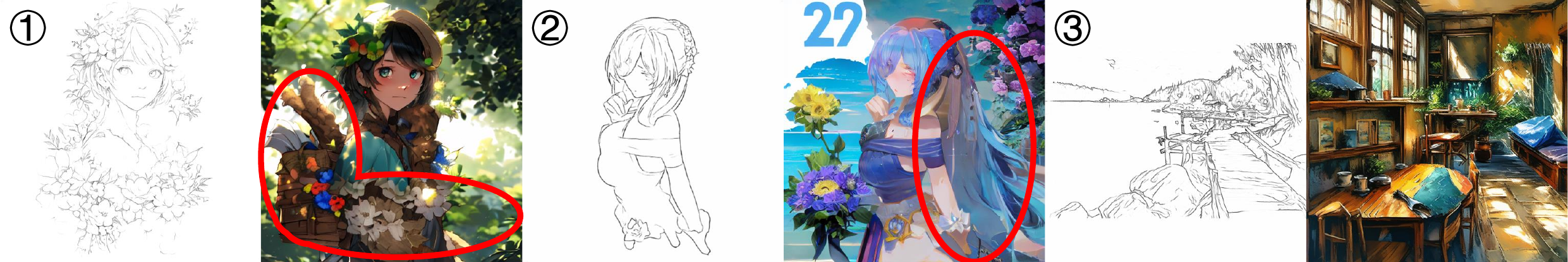}
    \caption{Examples of the distribution problem selected from Figure \ref{baseline}.}
    \label{highlight}
\end{figure}

We observed that with a higher GS and 0 drop rate, the \textit{Deform-0drop} and \textit{Shuffle-noisy} models achieved lower FID scores compared to the \textit{Shuffle-0drop} model, indicating that they perform better in terms of the quality of the generated images, possibly owing to the improvement of the distribution problem. The \textit{Dual} model achieved suboptimal FID scores compared to the other models, which we assume was due to the inappropriate $\lambda$ value in Eq. \ref{dual-loss}. However, considering the limitation of FID, which only quantifies the distance between the respective distributions of generated images and ground truth, we place greater emphasis on qualitative results for the distribution problem.\\

\noindent\textbf{Classifier-free Guidance and Drop Rate.} We estimated the generation performance of ablation models under different guidance scales, as shown in Table \ref{fid}. In order to observe the distribution problem, most of the models did not drop conditions during training. As shown in Figure \ref{cfg}, the \textit{Shuffle-0.8drop} model demonstrates better fidelity to the sketch input than the \textit{Shuffle-0drop} model under the same training epoch and sampling settings.

At the same time, the visually clear segmentation of results from the \textit{Shuffle-0drop} model under GS-1 and SGS-5 demonstrates that the network accurately recognizes faces. However, it exhibits a preference for synthesizing textures based on the reference, with its latent features located in $p(z|r)$. Increasing the reference drop rate can enhance the semantic fidelity to sketch inputs, but this effect tends to diminish as training progresses.\\

\noindent\textbf{Training Strategy and Training Epoch.} The training duration strongly affects the distribution problem, as illustrated in Figure \ref{actdistributions}, where the distribution $p_{\theta}(z|s,r)$ gradually shifts toward $p(z|r)$ as training progresses, observable in Figure \ref{sketch}. This shift occurs because the sketch conditions struggle to provide the semantics of fine-grained textures. The other qualitative evaluation of the training epoch is shown in Figure \ref{epoch}, where clear deterioration in segmentation can be observed in the results of the \textit{shuffle-0drop} model as it generated a human face. 

\begin{figure}[t]
	\centering
	\includegraphics[width=1\linewidth]{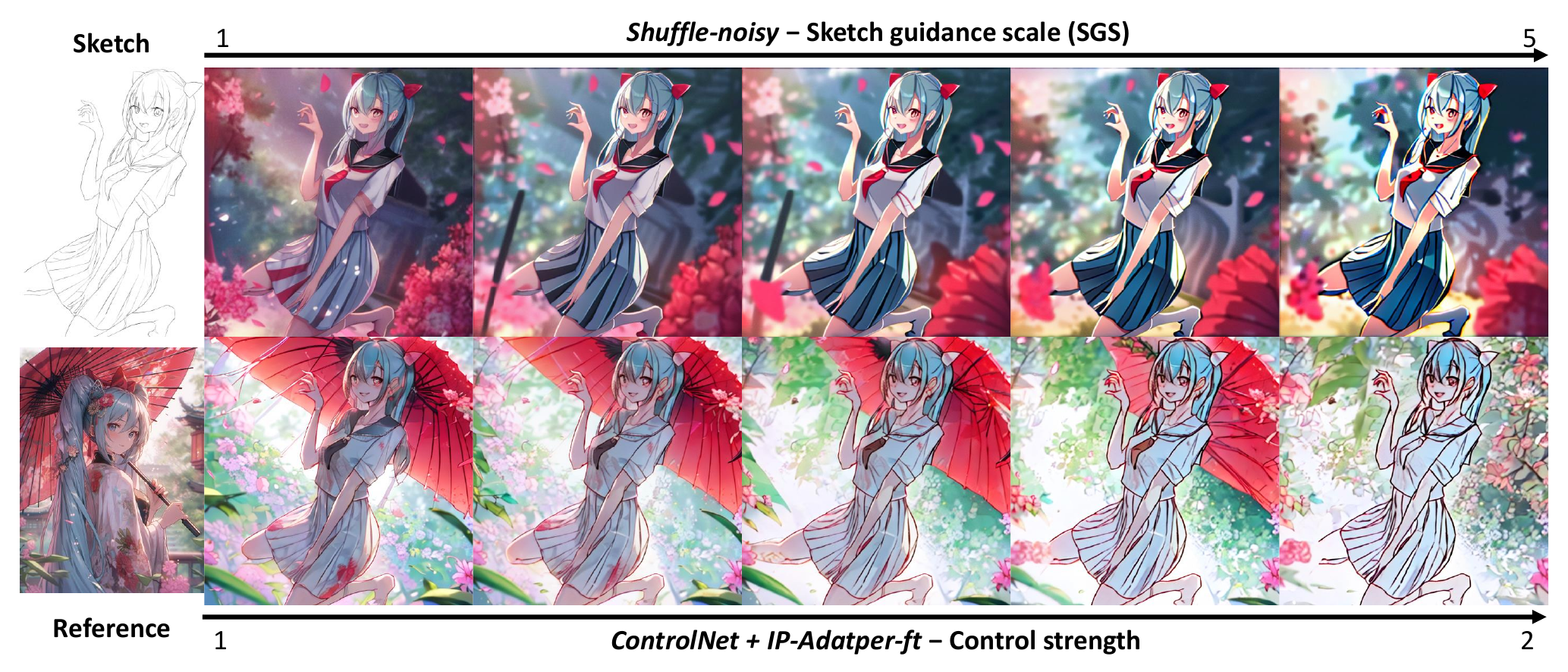}
	\caption{In contrast to the control scale used in \textit{ControlNet}, sketch-oriented CFG preserves the continuity of generated textures.}
	\label{fidelity}
\end{figure}
\begin{table}[t]
    \centering
    \caption{FID comparison between the \textit{Shuffle-noisy-7epoch} model and major baseline methods. We utilized Karras noise scheduler in this test \cite{KarrasAAL22}. Notably, the inferior quality of shuffled results suggests that T2I generation is also affected by the distribution problem. ``CN'': ControlNet; \dag: Texts were paired with unrelated sketch images.}
    \begin{tabular}{|c|c|c|c|c|c|}
        \hline
        \multicolumn{6}{|c|}{FID $\downarrow$}  \\
        \hline
         & GS-1 & GS-2 & GS-3 & GS-5 & GS-10 \\
        \hline
        \textit{Noisy-0} & 10.1036 & 11.1379 & 12.6028 & 14.4136 & 28.0530 \\
        \hline
		\multicolumn{6}{|c|}{Baseline}  \\
		\hline
		\multicolumn{5}{|c|}{\textit{CN-Anime\_Anything v3}, Text-based, GS-9} & 20.1411 \\
		\hline
		\multicolumn{5}{|c|}{\dag\textit{CN-Anime\_Anything v3}, Text-based, GS-9, Shuffle} & 27.4624 \\
  	\hline
		\multicolumn{5}{|c|}{\textit{CN-Lineart\_SD v1.5\_IP-Adapter}, GS-3} & 25.8390 \\
        \hline
        \multicolumn{5}{|c|}{\textit{CN-Anime\_Anything v3\_IP-Adapter-ft}, GS-3} & 23.2523 \\
		\hline
		\multicolumn{5}{|c|}{\textit{CN-Anime\_Anything v3\_IP-Adapter}, GS-3} & 39.2049 \\
		\hline
		\multicolumn{5}{|c|}{\textit{CN-Anime-Reference\_Anything v3}, GS-9} & 21.0125 \\
		\hline
        \multicolumn{5}{|c|}{\textit{CN-Canny-Anime\_SDXL\_IP-Adapter}, GS-3} &  35.8849 \\
		\hline
	\end{tabular}
    \label{baseline-fid}
\end{table}
\subsection{Comparison with Baseline}
We compare our method with baselines to validate the improvement achieved by decreasing the influence of the distribution problem. Considering the computational cost of training, we chose \textit{ControlNet} \cite{controlnet-iccv, controllllite, multi-controlnet, lvmin-injection}, \textit{IP-Adapter} \cite{ip-adapter, ip-adapter-hf}, and \textit{T2I-Adapter} \cite{t2i-adapter, t2i-adapter-code} as our major baselines. Most of them are publicly available, trained on large-scale datasets, and have demonstrated efficiency in generating high-quality images in various styles. Reference-based sketch colorization can be achieved by combining these adapters with a pre-trained SD model. We adopted three variations of SD in this evaluation: \textit{SD v1.5} \cite{RombachBLEO22,sd1.5-hf}, \textit{SDXL} \cite{sdxl,sdxl-hf}, and \textit{Anything v3} \cite{anything}. \textit{Anything v3} is a personalized SD model fine-tuned for generating anime-style images and is the backbone utilized to train the \textit{ControlNet-Anime} according to \cite{controlnet-v11}. 

Specifically, we fine-tuned the \textit{IP-Adapter v1.5} with \textit{Anything v3} on our training set for five epochs to align their distributions. The fine-tuned adapter is labeled as \textit{IP-Adapter-ft} in all experiments. The fine-tuned weight is included in our supplementary materials for validation.
\begin{figure}[t]
	\centering
	\includegraphics[width=1\linewidth]{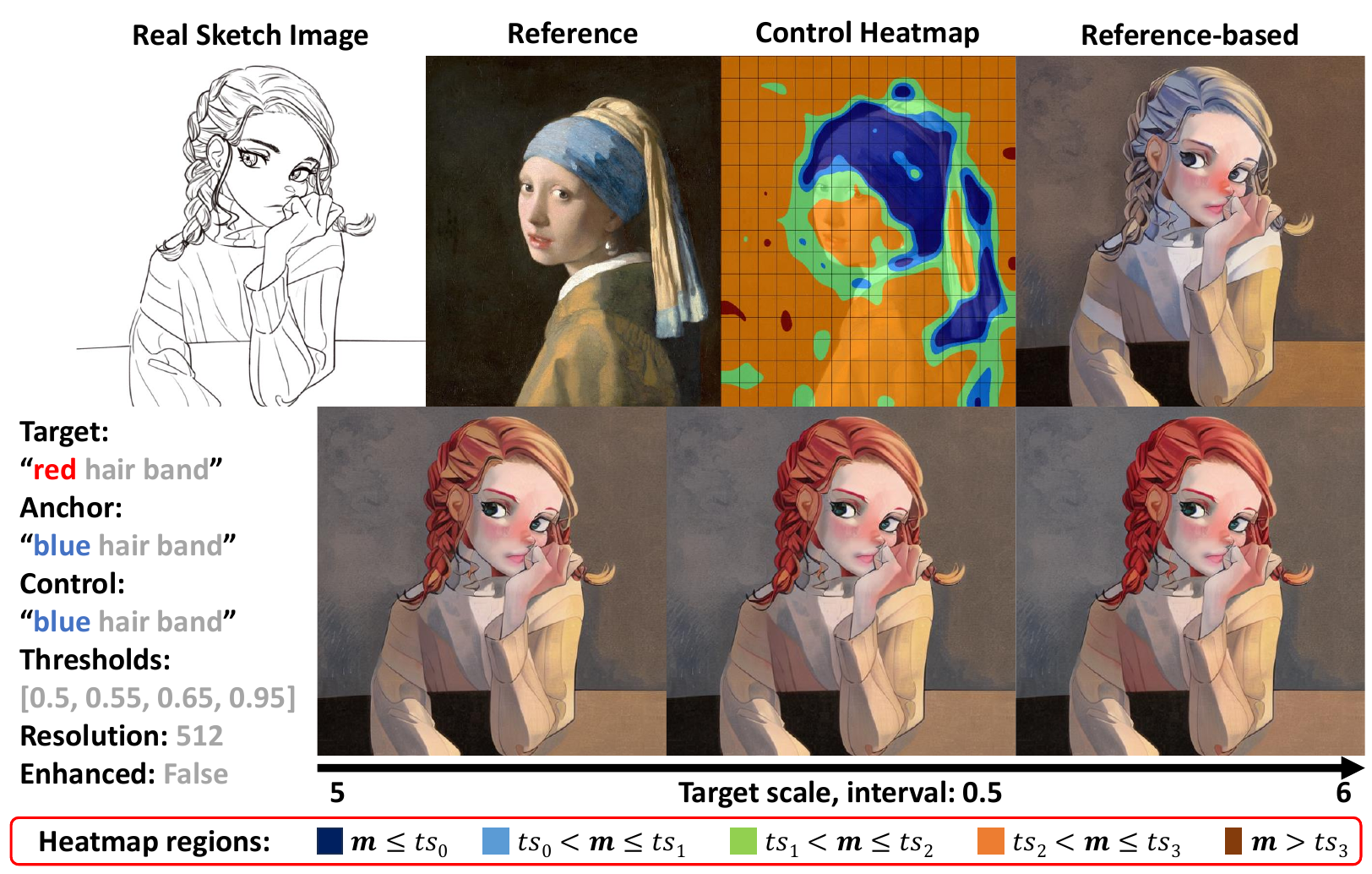}
	\caption{Visualization of the proposed local manipulation. The stratified heatmap displays the correlation vector $\bm{m}$ calculated on the basis of the control text.}
	\label{local-mani-vis-1}
\end{figure}
\begin{figure}[t]
	\centering
	\includegraphics[width=1\linewidth]{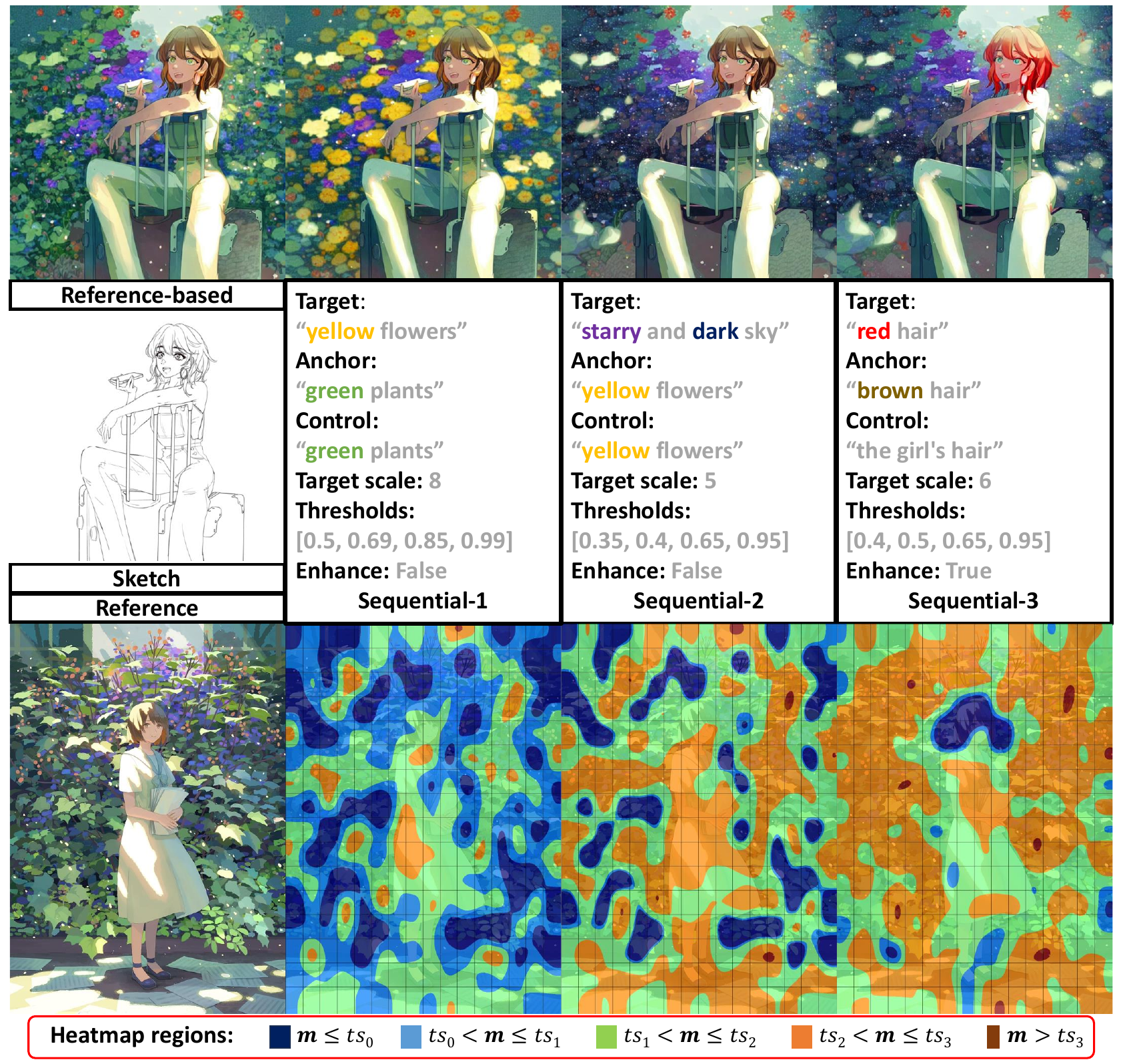}
	\caption{Illustration of the local manipulation performed sequentially.}
	\label{local-mani-vis-2}
\end{figure}
\begin{figure*}[t]
    \centering
    \includegraphics[width=1\linewidth]{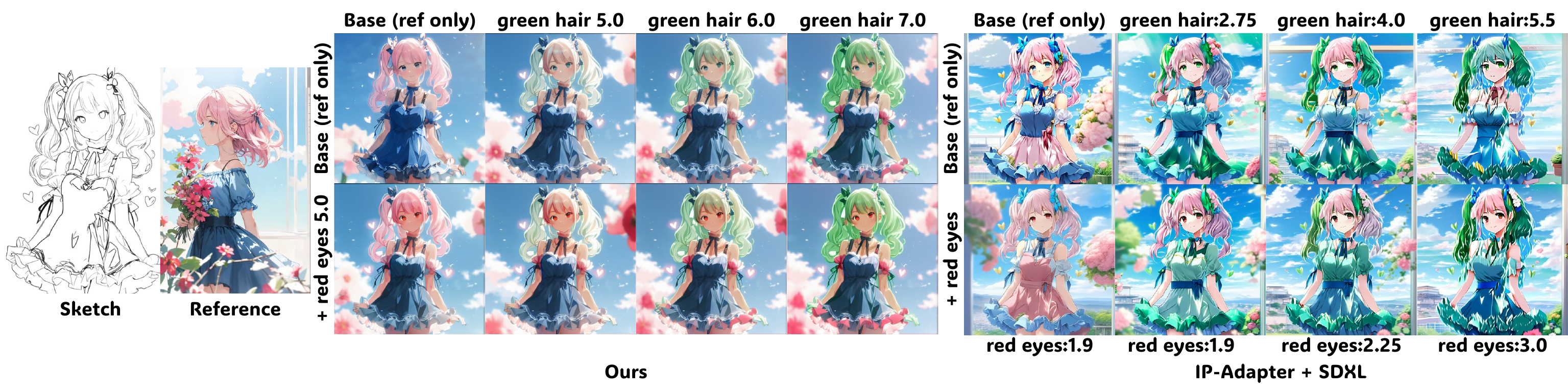}
    \caption{Comparison of text-based manipulation between our local manipulation and the combination of T2I-Adapter, SDXL, and ControlNet. When combined with image-guided adapters, SDXL tends to follow the guidance of text prompts less closely and needs higher weights if multiple attributes are jointly adjusted.}
    \label{comp-manipulation}
\end{figure*}
Necessary prompts were adopted for models originally designed for T2I generation, such as (``masterpiece, best quality, ultra-detailed, hires'') for positive prompts and (``easynegative'') \cite{easynegative} or (``negativeXL\_D'') \cite{negative-xl} for negative prompts. To avoid the distribution problem, we added ``a girl'' to the negative prompts when colorizing landscape sketch images with figure images, and to the positive prompts when using landscape images to colorize figure images.\\

\noindent\textbf{Quantitative Comparison.} Table \ref{baseline-fid} lists the FID scores of major baselines. For reference-based evaluation, color images were shuffled to colorize unrelated sketch images. The gap between the two text-based \textit{ControlNet} results is also notable, which highlights the considerable impact of the distribution problem on text-based generation.\\

\noindent\textbf{Qualitative Comparison.} As shown in Figure \ref{baseline}, our results typically feature better semantic fidelity to the sketch inputs and visually clearer segmentation compared to all baselines when applied to reference-based sketch colorization. Highlighted in Figure \ref{highlight}, where we can find many baseline methods changed the image composition and semantics of sketch inputs, some of which are highlighted in Figure \ref{highlight}: 1: Most of the flower sketches were ignored when rendering the bag. 2: Long hair was erroneously generated for the character. 3: The original semantics were destroyed. In contrast to the test in Figure \ref{distribution-vis}, for this comparison, we spent considerable time carefully adjusting the hyperparameters of the baseline methods to reduce the influence of the distribution problem on their results in rows (a)--(g). In contrast, we changed GS for our method, since the proposed models were trained using both conditions. 

We present the sketch-only T2I results in Figure \ref{baseline} to showcase the ideal composition of colorized results for comparison. Canny inputs, high-resolution images, and results generated using the default sampling settings of baseline methods are included in the supplementary materials.\\

\noindent\textbf{Sketch Fidelity.} Both our models and \textit{ControlNet} can increase the outputs' sketch fidelity using their respective hyperparameters, SGS and control strength. We here qualitatively compare their differences in a reference-based generation. As visualized in Figure \ref{fidelity}, the sketch-oriented CFG excels in maintaining color similarity with the original result (scale = 1) as the scale increases.

\subsection{Text-Based Manipulation}
\noindent\textbf{Global Manipulation.} Two qualitative experiments were conducted to evaluate the controllability of the \textit{CLS} model, where Figure \ref{teaserfigure} shows the results of our sequential global manipulation, which also demonstrates the effectiveness of progressive change. An example of detailed progressive manipulation is given in our supplementary materials.\\

\noindent\textbf{Local Manipulation.} Unlike global manipulation, which relies solely on the CLS token, local manipulation necessitates a PWV to adjust local tokens adaptively according to their association with the control text, leading to a more difficult manipulation. Figure \ref{local-mani-vis-1} demonstrates that local manipulation can progressively adjust a specific visual attribute, while Figure \ref{local-mani-vis-2} showcases sequential manipulation, altering backgrounds and hair color in sequential steps. Both figures adopt real sketch images. 

Although our method effectively adjusts visual attributes, a significant challenge arises from the proposed local manipulation. Observing the heatmaps in Figure \ref{local-mani-vis-2}, which were generated from projections on the control text embedding, reveals substantial errors in segmentation, which complicates the manipulation process.\\

\noindent\textbf{Compared to T2I Combination} T2I models can effectively adjust their colorized results when using only text prompts. However, when these models are combined with image prompts and additional adapters, the effectiveness of the text prompts may diminish. As shown in Figure \ref{comp-manipulation}, the text combination of \textit{SDXL}, \textit{ControlNet}, and \textit{IP-Adapter} is less likely to follow the guidance of text prompts.

\begin{figure}[t]
	\centering
	\includegraphics[width=1\linewidth]{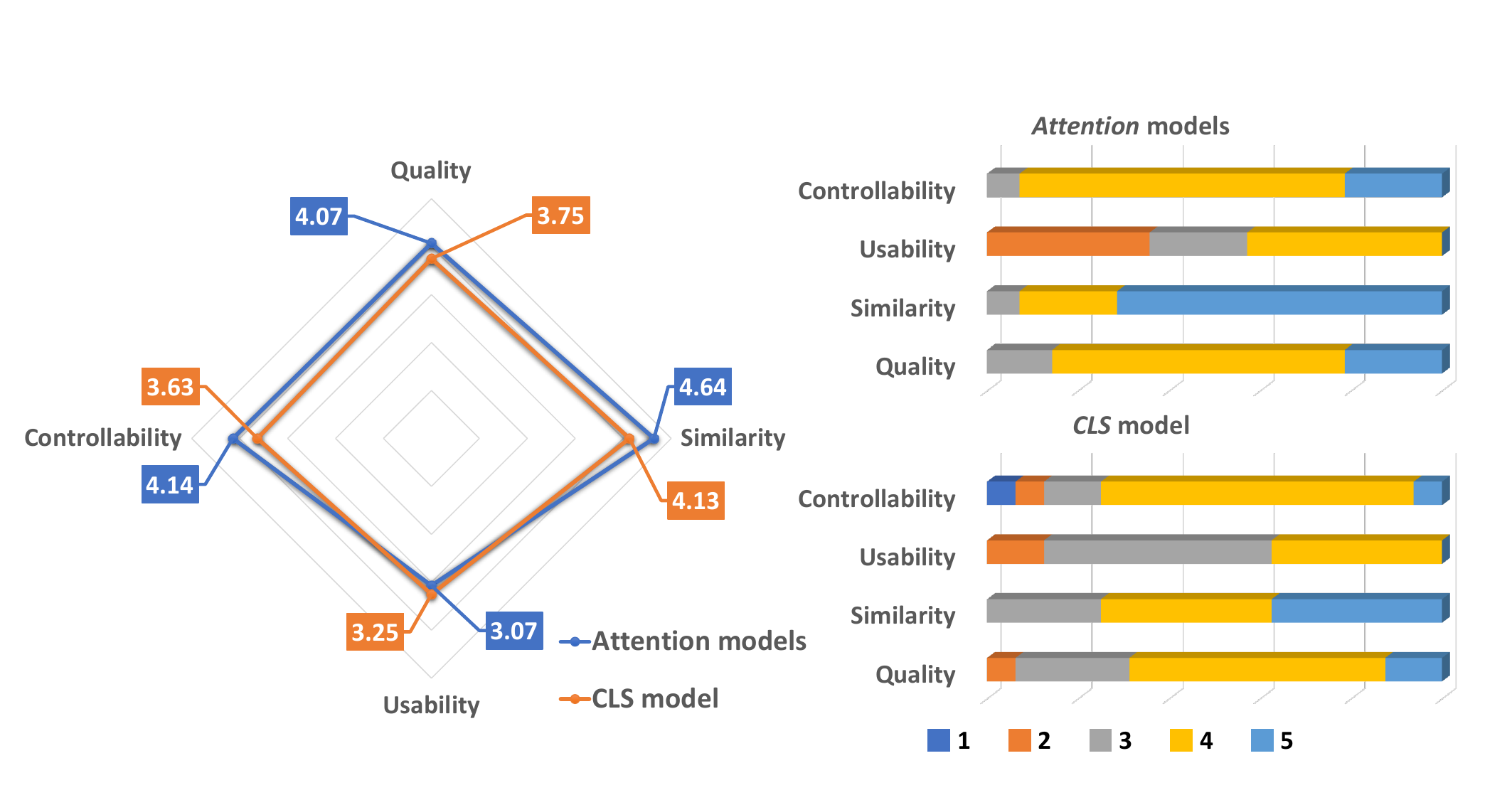}
        \caption{User study results. The radar charts show the average scores of four evaluations, and the bar chart showcases the distribution of user ratings.}
	\label{user-study}
\end{figure}
\subsection{User Study}
To evaluate our proposed methods subjectively, we implemented a user interface and invited 16 volunteers to experience our demo. Participants were required to test reference-based colorization and text-based manipulation for all proposed models. The average testing time for each individual exceeded one hour. After testing, we solicited participants' ratings across the following four dimensions. 
\begin{itemize}[leftmargin=*]
	\item[] \textbf{Quality}: Quality of generated images
	\item[] \textbf{Similarity}: Similarity with the reference image
	\item[] \textbf{Usability}: Ease of use
	\item[] \textbf{Controllability}: Correspondence between manipulated results and target texts
\end{itemize}
The results, as shown in Figure \ref{user-study}, indicate overall satisfaction with image quality, control, and similarity. However, the relatively lower usability score demonstrates that the proposed manipulation requires further refinement to achieve simplicity.
	\section{conclusion}
In this paper, we presented a thorough examination of the application of reference-based SD to sketch colorization. We analyzed how the distribution problem leads to inferior outputs compared to text-based models and offered a general solution to diminish its impact. Leveraging a pre-trained CLIP, we proposed two variations of reference-based colorization SD and two kinds of zero-shot sequential manipulation methods. Our experimental results, including qualitative/quantitative evaluations and user studies, validate the effectiveness of our reference-based colorization and text-based manipulation methods.
However, our work has four primary limitations, as follows.
\begin{itemize}[leftmargin=*]
	\item[1.] Achieving precise segmentation based solely on the control text is challenging in the proposed local manipulation. In addition, manipulation without self-adaptive trainable modules struggles to replicate the real changes of tokens, especially for high-level embeddings determined by all tokens, such as ``daytime'' and ``night''.
	\item[2.] Because our manipulation is based on image prompts, it is inevitable that some semantically unrelated visual attributes will be changed because they are colorized based on the manipulated regions in the reference. This can be observed in Figure \ref{local-mani-vis-2}, where the color of the right suitcase is changed.
	\item[3.] Since our models were trained for high-fidelity sketch colorization, they are unsuitable for inpainting if the edge of the sketch is too sharp, which is observable in rows (f) and (j) in Figure \ref{baseline}.
        \item[4.] The proposed solutions to the distribution problems are trade-off methods, which result in less fine-grained textures and simple backgrounds when given rough sketches due to the characteristic of features in $p_{\theta}(z|s)$.
\end{itemize}
Our future work will primarily focus on proposing improved methods and well-designed architectures to further eliminate the distribution problem. We will also work on designing a metric to evaluate the distribution problem quantitatively and enhancing the usability and controllability of local manipulation through three potential methods: 1) introducing a trainable module for adaptive PWV computation, 2) directly modifying features during the denoising process, and 3) designing advanced interactive systems to assist users in the selection of regions for local manipulation.
\bibliographystyle{ACM-Reference-Format} 
\bibliography{sample-base}


\begin{thebibliography}{79}

%%% the Web address.

\ifx \showCODEN    \undefined \def \showCODEN     #1{\unskip}     \fi
\ifx \showDOI      \undefined \def \showDOI       #1{#1}\fi
\ifx \showISBNx    \undefined \def \showISBNx     #1{\unskip}     \fi
\ifx \showISBNxiii \undefined \def \showISBNxiii  #1{\unskip}     \fi
\ifx \showISSN     \undefined \def \showISSN      #1{\unskip}     \fi
\ifx \showLCCN     \undefined \def \showLCCN      #1{\unskip}     \fi
\ifx \shownote     \undefined \def \shownote      #1{#1}          \fi
\ifx \showarticletitle \undefined \def \showarticletitle #1{#1}   \fi
\ifx \showURL      \undefined \def \showURL       {\relax}        \fi
% invisible to TeX
\providecommand\bibfield[2]{#2}
\providecommand\bibinfo[2]{#2}
\providecommand\natexlab[1]{#1}
\providecommand\showeprint[2][]{arXiv:#2}

\bibitem[Akita et~al\mbox{.}(2020)]%
        {AkitaMT20}
\bibfield{author}{\bibinfo{person}{Kenta Akita}, \bibinfo{person}{Yuki Morimoto}, {and} \bibinfo{person}{Reiji Tsuruno}.} \bibinfo{year}{2020}\natexlab{}.
\newblock \showarticletitle{Colorization of Line Drawings with Empty Pupils}.
\newblock \bibinfo{journal}{\emph{Comput. Graph. Forum}} \bibinfo{volume}{39}, \bibinfo{number}{7} (\bibinfo{year}{2020}), \bibinfo{pages}{601--610}.
\newblock
\urldef\tempurl%
\url{https://doi.org/10.1111/cgf.14171}
\showDOI{\tempurl}


\bibitem[Cao et~al\mbox{.}(2023)]%
        {abs-2303-11137}
\bibfield{author}{\bibinfo{person}{Yu Cao}, \bibinfo{person}{Xiangqiao Meng}, \bibinfo{person}{P.~Y. Mok}, \bibinfo{person}{Xueting Liu}, \bibinfo{person}{Tong{-}Yee Lee}, {and} \bibinfo{person}{Ping Li}.} \bibinfo{year}{2023}\natexlab{}.
\newblock \showarticletitle{AnimeDiffusion: Anime Face Line Drawing Colorization via Diffusion Models}.
\newblock \bibinfo{journal}{\emph{CoRR}}  \bibinfo{volume}{abs/2303.11137} (\bibinfo{year}{2023}).
\newblock
\urldef\tempurl%
\url{https://doi.org/10.48550/ARXIV.2303.11137}
\showDOI{\tempurl}


\bibitem[Cherti et~al\mbox{.}(2023)]%
        {openclip-2}
\bibfield{author}{\bibinfo{person}{Mehdi Cherti}, \bibinfo{person}{Romain Beaumont}, \bibinfo{person}{Ross Wightman}, \bibinfo{person}{Mitchell Wortsman}, \bibinfo{person}{Gabriel Ilharco}, \bibinfo{person}{Cade Gordon}, \bibinfo{person}{Christoph Schuhmann}, \bibinfo{person}{Ludwig Schmidt}, {and} \bibinfo{person}{Jenia Jitsev}.} \bibinfo{year}{2023}\natexlab{}.
\newblock \showarticletitle{Reproducible scaling laws for contrastive language-image learning}. In \bibinfo{booktitle}{\emph{{CVPR}}}. \bibinfo{pages}{2818--2829}.
\newblock


\bibitem[Choi et~al\mbox{.}(2018)]%
        {ChoiCKH0C18}
\bibfield{author}{\bibinfo{person}{Yunjey Choi}, \bibinfo{person}{Min{-}Je Choi}, \bibinfo{person}{Munyoung Kim}, \bibinfo{person}{Jung{-}Woo Ha}, \bibinfo{person}{Sunghun Kim}, {and} \bibinfo{person}{Jaegul Choo}.} \bibinfo{year}{2018}\natexlab{}.
\newblock \showarticletitle{StarGAN: Unified Generative Adversarial Networks for Multi-Domain Image-to-Image Translation}. In \bibinfo{booktitle}{\emph{{CVPR}}}. \bibinfo{publisher}{{IEEE/CVF}}, \bibinfo{pages}{8789--8797}.
\newblock
\urldef\tempurl%
\url{https://doi.org/10.1109/CVPR.2018.00916}
\showDOI{\tempurl}


\bibitem[Choi et~al\mbox{.}(2020)]%
        {ChoiUYH20}
\bibfield{author}{\bibinfo{person}{Yunjey Choi}, \bibinfo{person}{Youngjung Uh}, \bibinfo{person}{Jaejun Yoo}, {and} \bibinfo{person}{Jung{-}Woo Ha}.} \bibinfo{year}{2020}\natexlab{}.
\newblock \showarticletitle{StarGAN v2: Diverse Image Synthesis for Multiple Domains}. In \bibinfo{booktitle}{\emph{{CVPR}}}. \bibinfo{publisher}{{IEEE/CVF}}, \bibinfo{pages}{8185--8194}.
\newblock
\urldef\tempurl%
\url{https://doi.org/10.1109/CVPR42600.2020.00821}
\showDOI{\tempurl}


\bibitem[community et~al\mbox{.}(2022)]%
        {danbooru2021}
\bibfield{author}{\bibinfo{person}{Danbooru community}, \bibinfo{person}{Gwern Branwen}, {and} \bibinfo{person}{Anonymous}.} \bibinfo{year}{2022}\natexlab{}.
\newblock \bibinfo{title}{Danbooru2021: A Large-Scale Crowdsourced and Tagged Anime Illustration Dataset}.
\newblock \bibinfo{howpublished}{\url{https://gwern.net/danbooru2021}}.
\newblock
\newblock
\shownote{Accessed: DATE 2022-01-21}.


\bibitem[Dhariwal and Nichol(2021)]%
        {DhariwalN21}
\bibfield{author}{\bibinfo{person}{Prafulla Dhariwal} {and} \bibinfo{person}{Alexander~Quinn Nichol}.} \bibinfo{year}{2021}\natexlab{}.
\newblock \showarticletitle{Diffusion Models Beat GANs on Image Synthesis}. In \bibinfo{booktitle}{\emph{{NeurIPS}}}. \bibinfo{pages}{8780--8794}.
\newblock


\bibitem[Esser et~al\mbox{.}(2021)]%
        {EsserRO21}
\bibfield{author}{\bibinfo{person}{Patrick Esser}, \bibinfo{person}{Robin Rombach}, {and} \bibinfo{person}{Bj{\"{o}}rn Ommer}.} \bibinfo{year}{2021}\natexlab{}.
\newblock \showarticletitle{Taming Transformers for High-Resolution Image Synthesis}. In \bibinfo{booktitle}{\emph{{CVPR}}}. \bibinfo{publisher}{{IEEE/CVF}}, \bibinfo{pages}{12873--12883}.
\newblock
\urldef\tempurl%
\url{https://doi.org/10.1109/CVPR46437.2021.01268}
\showDOI{\tempurl}


\bibitem[Fourey et~al\mbox{.}(2018)]%
        {RevoyFT18}
\bibfield{author}{\bibinfo{person}{S{\'{e}}bastien Fourey}, \bibinfo{person}{David Tschumperl{\'{e}}}, {and} \bibinfo{person}{David Revoy}.} \bibinfo{year}{2018}\natexlab{}.
\newblock \showarticletitle{A Fast and Efficient Semi-guided Algorithm for Flat Coloring Line-arts}. In \bibinfo{booktitle}{\emph{Vision, Modeling and Visualization {VMV}}}. \bibinfo{publisher}{Eurographics Association}, \bibinfo{pages}{1--9}.
\newblock
\urldef\tempurl%
\url{https://doi.org/10.2312/vmv.20181247}
\showDOI{\tempurl}


\bibitem[Furusawa et~al\mbox{.}(2017)]%
        {FurusawaHOO17}
\bibfield{author}{\bibinfo{person}{Chie Furusawa}, \bibinfo{person}{Kazuyuki Hiroshiba}, \bibinfo{person}{Keisuke Ogaki}, {and} \bibinfo{person}{Yuri Odagiri}.} \bibinfo{year}{2017}\natexlab{}.
\newblock \showarticletitle{Comicolorization: semi-automatic manga colorization}. In \bibinfo{booktitle}{\emph{{SIGGRAPH Asia}}}. \bibinfo{publisher}{{ACM}}, \bibinfo{pages}{12:1--12:4}.
\newblock
\urldef\tempurl%
\url{https://doi.org/10.1145/3145749.3149430}
\showDOI{\tempurl}


\bibitem[Gal et~al\mbox{.}(2022)]%
        {GalPMBCC22}
\bibfield{author}{\bibinfo{person}{Rinon Gal}, \bibinfo{person}{Or Patashnik}, \bibinfo{person}{Haggai Maron}, \bibinfo{person}{Amit~H. Bermano}, \bibinfo{person}{Gal Chechik}, {and} \bibinfo{person}{Daniel Cohen{-}Or}.} \bibinfo{year}{2022}\natexlab{}.
\newblock \showarticletitle{StyleGAN-NADA: CLIP-guided domain adaptation of image generators}.
\newblock \bibinfo{journal}{\emph{{ACM} Trans. Graph.}} \bibinfo{volume}{41}, \bibinfo{number}{4} (\bibinfo{year}{2022}), \bibinfo{pages}{141:1--141:13}.
\newblock
\urldef\tempurl%
\url{https://doi.org/10.1145/3528223.3530164}
\showDOI{\tempurl}


\bibitem[Gatys et~al\mbox{.}(2016)]%
        {GatysEB16}
\bibfield{author}{\bibinfo{person}{Leon~A. Gatys}, \bibinfo{person}{Alexander~S. Ecker}, {and} \bibinfo{person}{Matthias Bethge}.} \bibinfo{year}{2016}\natexlab{}.
\newblock \showarticletitle{Image Style Transfer Using Convolutional Neural Networks}. In \bibinfo{booktitle}{\emph{{CVPR}}}. \bibinfo{publisher}{{IEEE/CVF}}, \bibinfo{pages}{2414--2423}.
\newblock
\urldef\tempurl%
\url{https://doi.org/10.1109/CVPR.2016.265}
\showDOI{\tempurl}


\bibitem[Goodfellow et~al\mbox{.}(2014)]%
        {GoodfellowPMXWOCB14}
\bibfield{author}{\bibinfo{person}{Ian~J. Goodfellow}, \bibinfo{person}{Jean Pouget{-}Abadie}, \bibinfo{person}{Mehdi Mirza}, \bibinfo{person}{Bing Xu}, \bibinfo{person}{David Warde{-}Farley}, \bibinfo{person}{Sherjil Ozair}, \bibinfo{person}{Aaron~C. Courville}, {and} \bibinfo{person}{Yoshua Bengio}.} \bibinfo{year}{2014}\natexlab{}.
\newblock \showarticletitle{Generative Adversarial Nets}. In \bibinfo{booktitle}{\emph{{NeurIPS}}}. \bibinfo{pages}{2672--2680}.
\newblock


\bibitem[h94(2024)]%
        {ip-adapter-hf}
\bibfield{author}{\bibinfo{person}{h94}.} \bibinfo{year}{2024}\natexlab{}.
\newblock \bibinfo{title}{Hugging Face/IP-Adapter}.
\newblock \bibinfo{howpublished}{\url{https://huggingface.co/h94/IP-Adapter}}.
\newblock
\newblock
\shownote{Accessed: DATE 2024-01-02}.


\bibitem[Hakurei(2023)]%
        {waifudiffusion}
\bibfield{author}{\bibinfo{person}{Reimu Hakurei}.} \bibinfo{year}{2023}\natexlab{}.
\newblock \bibinfo{title}{Hugging Face/waifu-diffusion-v1-4}.
\newblock \bibinfo{howpublished}{\url{https://huggingface.co/hakurei/waifu-diffusion-v1-4}}.
\newblock
\newblock
\shownote{Accessed: DATE 2023-03-05}.


\bibitem[Havoc(2023)]%
        {easynegative}
\bibfield{author}{\bibinfo{person}{Havoc}.} \bibinfo{year}{2023}\natexlab{}.
\newblock \bibinfo{title}{EasyNegative}.
\newblock \bibinfo{howpublished}{\url{https://civitai.com/models/7808/easynegative}}.
\newblock
\newblock
\shownote{Accessed: DATE 2023-02-10}.


\bibitem[He et~al\mbox{.}(2018)]%
        {he2018deep}
\bibfield{author}{\bibinfo{person}{Mingming He}, \bibinfo{person}{Dongdong Chen}, \bibinfo{person}{Jing Liao}, \bibinfo{person}{Pedro~V Sander}, {and} \bibinfo{person}{Lu Yuan}.} \bibinfo{year}{2018}\natexlab{}.
\newblock \showarticletitle{Deep exemplar-based colorization}.
\newblock \bibinfo{journal}{\emph{{ACM} Trans. Graph.}} \bibinfo{volume}{37}, \bibinfo{number}{4} (\bibinfo{year}{2018}), \bibinfo{pages}{47}.
\newblock


\bibitem[Heusel et~al\mbox{.}(2017)]%
        {HeuselRUNH17}
\bibfield{author}{\bibinfo{person}{Martin Heusel}, \bibinfo{person}{Hubert Ramsauer}, \bibinfo{person}{Thomas Unterthiner}, \bibinfo{person}{Bernhard Nessler}, {and} \bibinfo{person}{Sepp Hochreiter}.} \bibinfo{year}{2017}\natexlab{}.
\newblock \showarticletitle{GANs Trained by a Two Time-Scale Update Rule Converge to a Local Nash Equilibrium}. In \bibinfo{booktitle}{\emph{{NeurIPS}}}. \bibinfo{pages}{6626--6637}.
\newblock


\bibitem[Ho et~al\mbox{.}(2020)]%
        {HoJA20}
\bibfield{author}{\bibinfo{person}{Jonathan Ho}, \bibinfo{person}{Ajay Jain}, {and} \bibinfo{person}{Pieter Abbeel}.} \bibinfo{year}{2020}\natexlab{}.
\newblock \showarticletitle{Denoising Diffusion Probabilistic Models}. In \bibinfo{booktitle}{\emph{{NeurIPS}}}.
\newblock


\bibitem[Ho and Salimans(2022)]%
        {abs-2207-12598}
\bibfield{author}{\bibinfo{person}{Jonathan Ho} {and} \bibinfo{person}{Tim Salimans}.} \bibinfo{year}{2022}\natexlab{}.
\newblock \showarticletitle{Classifier-Free Diffusion Guidance}.
\newblock \bibinfo{journal}{\emph{CoRR}}  \bibinfo{volume}{abs/2207.12598} (\bibinfo{year}{2022}).
\newblock
\urldef\tempurl%
\url{https://doi.org/10.48550/arXiv.2207.12598}
\showDOI{\tempurl}


\bibitem[Hu et~al\mbox{.}(2022)]%
        {HuSWALWWC22}
\bibfield{author}{\bibinfo{person}{Edward~J. Hu}, \bibinfo{person}{Yelong Shen}, \bibinfo{person}{Phillip Wallis}, \bibinfo{person}{Zeyuan Allen{-}Zhu}, \bibinfo{person}{Yuanzhi Li}, \bibinfo{person}{Shean Wang}, \bibinfo{person}{Lu Wang}, {and} \bibinfo{person}{Weizhu Chen}.} \bibinfo{year}{2022}\natexlab{}.
\newblock \showarticletitle{LoRA: Low-Rank Adaptation of Large Language Models}. In \bibinfo{booktitle}{\emph{{ICLR}}}. \bibinfo{publisher}{OpenReview.net}.
\newblock


\bibitem[Huang and Belongie(2017)]%
        {HuangB17}
\bibfield{author}{\bibinfo{person}{Xun Huang} {and} \bibinfo{person}{Serge~J. Belongie}.} \bibinfo{year}{2017}\natexlab{}.
\newblock \showarticletitle{Arbitrary Style Transfer in Real-Time with Adaptive Instance Normalization}. In \bibinfo{booktitle}{\emph{{ICCV}}}. \bibinfo{publisher}{{IEEE/CVF}}, \bibinfo{pages}{1510--1519}.
\newblock
\urldef\tempurl%
\url{https://doi.org/10.1109/ICCV.2017.167}
\showDOI{\tempurl}


\bibitem[Ilharco et~al\mbox{.}(2021)]%
        {openclip}
\bibfield{author}{\bibinfo{person}{Gabriel Ilharco}, \bibinfo{person}{Mitchell Wortsman}, \bibinfo{person}{Ross Wightman}, \bibinfo{person}{Cade Gordon}, \bibinfo{person}{Nicholas Carlini}, \bibinfo{person}{Rohan Taori}, \bibinfo{person}{Achal Dave}, \bibinfo{person}{Vaishaal Shankar}, \bibinfo{person}{Hongseok Namkoong}, \bibinfo{person}{John Miller}, \bibinfo{person}{Hannaneh Hajishirzi}, \bibinfo{person}{Ali Farhadi}, {and} \bibinfo{person}{Ludwig Schmidt}.} \bibinfo{year}{2021}\natexlab{}.
\newblock \bibinfo{booktitle}{\emph{OpenCLIP}}.
\newblock
\urldef\tempurl%
\url{https://doi.org/10.5281/zenodo.5143773}
\showDOI{\tempurl}


\bibitem[Isola et~al\mbox{.}(2017)]%
        {IsolaZZE17}
\bibfield{author}{\bibinfo{person}{Phillip Isola}, \bibinfo{person}{Jun{-}Yan Zhu}, \bibinfo{person}{Tinghui Zhou}, {and} \bibinfo{person}{Alexei~A. Efros}.} \bibinfo{year}{2017}\natexlab{}.
\newblock \showarticletitle{Image-to-Image Translation with Conditional Adversarial Networks}. In \bibinfo{booktitle}{\emph{{CVPR}}}. \bibinfo{publisher}{{IEEE/CVF}}, \bibinfo{pages}{5967--5976}.
\newblock
\urldef\tempurl%
\url{https://doi.org/10.1109/CVPR.2017.632}
\showDOI{\tempurl}


\bibitem[Johnson et~al\mbox{.}(2016)]%
        {JohnsonAF16}
\bibfield{author}{\bibinfo{person}{Justin Johnson}, \bibinfo{person}{Alexandre Alahi}, {and} \bibinfo{person}{Li Fei{-}Fei}.} \bibinfo{year}{2016}\natexlab{}.
\newblock \showarticletitle{Perceptual Losses for Real-Time Style Transfer and Super-Resolution}. In \bibinfo{booktitle}{\emph{{ECCV}}}, Vol.~\bibinfo{volume}{9906}. \bibinfo{publisher}{Springer}, \bibinfo{pages}{694--711}.
\newblock
\urldef\tempurl%
\url{https://doi.org/10.1007/978-3-319-46475-6\_43}
\showDOI{\tempurl}


\bibitem[Karras et~al\mbox{.}(2022)]%
        {KarrasAAL22}
\bibfield{author}{\bibinfo{person}{Tero Karras}, \bibinfo{person}{Miika Aittala}, \bibinfo{person}{Timo Aila}, {and} \bibinfo{person}{Samuli Laine}.} \bibinfo{year}{2022}\natexlab{}.
\newblock \showarticletitle{Elucidating the Design Space of Diffusion-Based Generative Models}. In \bibinfo{booktitle}{\emph{{NeurIPS}}}, \bibfield{editor}{\bibinfo{person}{Sanmi Koyejo}, \bibinfo{person}{S.~Mohamed}, \bibinfo{person}{A.~Agarwal}, \bibinfo{person}{Danielle Belgrave}, \bibinfo{person}{K.~Cho}, {and} \bibinfo{person}{A.~Oh}} (Eds.).
\newblock


\bibitem[Karras et~al\mbox{.}(2019)]%
        {KarrasLA19}
\bibfield{author}{\bibinfo{person}{Tero Karras}, \bibinfo{person}{Samuli Laine}, {and} \bibinfo{person}{Timo Aila}.} \bibinfo{year}{2019}\natexlab{}.
\newblock \showarticletitle{A Style-Based Generator Architecture for Generative Adversarial Networks}. In \bibinfo{booktitle}{\emph{{CVPR}}}. \bibinfo{publisher}{{IEEE/CVF}}, \bibinfo{pages}{4401--4410}.
\newblock
\urldef\tempurl%
\url{https://doi.org/10.1109/CVPR.2019.00453}
\showDOI{\tempurl}


\bibitem[Karras et~al\mbox{.}(2020)]%
        {KarrasLAHLA20}
\bibfield{author}{\bibinfo{person}{Tero Karras}, \bibinfo{person}{Samuli Laine}, \bibinfo{person}{Miika Aittala}, \bibinfo{person}{Janne Hellsten}, \bibinfo{person}{Jaakko Lehtinen}, {and} \bibinfo{person}{Timo Aila}.} \bibinfo{year}{2020}\natexlab{}.
\newblock \showarticletitle{Analyzing and Improving the Image Quality of StyleGAN}. In \bibinfo{booktitle}{\emph{{CVPR}}}. \bibinfo{publisher}{{IEEE/CVF}}, \bibinfo{pages}{8107--8116}.
\newblock
\urldef\tempurl%
\url{https://doi.org/10.1109/CVPR42600.2020.00813}
\showDOI{\tempurl}


\bibitem[Kim et~al\mbox{.}(2022)]%
        {KimKY22a}
\bibfield{author}{\bibinfo{person}{Gwanghyun Kim}, \bibinfo{person}{Taesung Kwon}, {and} \bibinfo{person}{Jong~Chul Ye}.} \bibinfo{year}{2022}\natexlab{}.
\newblock \showarticletitle{DiffusionCLIP: Text-Guided Diffusion Models for Robust Image Manipulation}. In \bibinfo{booktitle}{\emph{{CVPR}}}. \bibinfo{publisher}{{IEEE/CVF}}, \bibinfo{pages}{2416--2425}.
\newblock
\urldef\tempurl%
\url{https://doi.org/10.1109/CVPR52688.2022.00246}
\showDOI{\tempurl}


\bibitem[Kim et~al\mbox{.}(2019)]%
        {KimJPY19}
\bibfield{author}{\bibinfo{person}{Hyunsu Kim}, \bibinfo{person}{Ho~Young Jhoo}, \bibinfo{person}{Eunhyeok Park}, {and} \bibinfo{person}{Sungjoo Yoo}.} \bibinfo{year}{2019}\natexlab{}.
\newblock \showarticletitle{Tag2Pix: Line Art Colorization Using Text Tag With SECat and Changing Loss}. In \bibinfo{booktitle}{\emph{{ICCV}}}. \bibinfo{publisher}{{IEEE/CVF}}, \bibinfo{pages}{9055--9064}.
\newblock
\urldef\tempurl%
\url{https://doi.org/10.1109/ICCV.2019.00915}
\showDOI{\tempurl}


\bibitem[Kingma and Ba(2015)]%
        {KingmaB14}
\bibfield{author}{\bibinfo{person}{Diederik~P. Kingma} {and} \bibinfo{person}{Jimmy Ba}.} \bibinfo{year}{2015}\natexlab{}.
\newblock \showarticletitle{Adam: {A} Method for Stochastic Optimization}. In \bibinfo{booktitle}{\emph{{ICLR}}}.
\newblock


\bibitem[Kingma and Welling(2014)]%
        {KingmaW13}
\bibfield{author}{\bibinfo{person}{Diederik~P. Kingma} {and} \bibinfo{person}{Max Welling}.} \bibinfo{year}{2014}\natexlab{}.
\newblock \showarticletitle{Auto-Encoding Variational Bayes}. In \bibinfo{booktitle}{\emph{{ICLR}}}.
\newblock


\bibitem[kohya ss(2024)]%
        {controllllite}
\bibfield{author}{\bibinfo{person}{kohya ss}.} \bibinfo{year}{2024}\natexlab{}.
\newblock \bibinfo{title}{Hugging Face/controlnet-lllite}.
\newblock \bibinfo{howpublished}{\url{https://huggingface.co/kohya-ss/controlnet-lllite}}.
\newblock
\newblock
\shownote{Accessed: DATE 2024-01-02}.


\bibitem[Lee et~al\mbox{.}(2020)]%
        {LeeKLKCC20}
\bibfield{author}{\bibinfo{person}{Junsoo Lee}, \bibinfo{person}{Eungyeup Kim}, \bibinfo{person}{Yunsung Lee}, \bibinfo{person}{Dongjun Kim}, \bibinfo{person}{Jaehyuk Chang}, {and} \bibinfo{person}{Jaegul Choo}.} \bibinfo{year}{2020}\natexlab{}.
\newblock \showarticletitle{Reference-Based Sketch Image Colorization Using Augmented-Self Reference and Dense Semantic Correspondence}. In \bibinfo{booktitle}{\emph{{CVPR}}}. \bibinfo{publisher}{{IEEE/CVF}}, \bibinfo{pages}{5800--5809}.
\newblock
\urldef\tempurl%
\url{https://doi.org/10.1109/CVPR42600.2020.00584}
\showDOI{\tempurl}


\bibitem[Li et~al\mbox{.}(2022)]%
        {li2022eliminating}
\bibfield{author}{\bibinfo{person}{Zekun Li}, \bibinfo{person}{Zhengyang Geng}, \bibinfo{person}{Zhao Kang}, \bibinfo{person}{Wenyu Chen}, {and} \bibinfo{person}{Yibo Yang}.} \bibinfo{year}{2022}\natexlab{}.
\newblock \showarticletitle{Eliminating Gradient Conflict in Reference-based Line-Art Colorization}. In \bibinfo{booktitle}{\emph{{ECCV}}}. \bibinfo{publisher}{Springer}, \bibinfo{pages}{579--596}.
\newblock


\bibitem[Liu et~al\mbox{.}(2023)]%
        {LiuPAZCHSRD23}
\bibfield{author}{\bibinfo{person}{Xihui Liu}, \bibinfo{person}{Dong~Huk Park}, \bibinfo{person}{Samaneh Azadi}, \bibinfo{person}{Gong Zhang}, \bibinfo{person}{Arman Chopikyan}, \bibinfo{person}{Yuxiao Hu}, \bibinfo{person}{Humphrey Shi}, \bibinfo{person}{Anna Rohrbach}, {and} \bibinfo{person}{Trevor Darrell}.} \bibinfo{year}{2023}\natexlab{}.
\newblock \showarticletitle{More Control for Free! Image Synthesis with Semantic Diffusion Guidance}. In \bibinfo{booktitle}{\emph{{WACV}}}. \bibinfo{publisher}{{IEEE/CVF}}, \bibinfo{pages}{289--299}.
\newblock
\urldef\tempurl%
\url{https://doi.org/10.1109/WACV56688.2023.00037}
\showDOI{\tempurl}


\bibitem[Loshchilov and Hutter(2019)]%
        {LoshchilovH19}
\bibfield{author}{\bibinfo{person}{Ilya Loshchilov} {and} \bibinfo{person}{Frank Hutter}.} \bibinfo{year}{2019}\natexlab{}.
\newblock \showarticletitle{Decoupled Weight Decay Regularization}. In \bibinfo{booktitle}{\emph{{ICLR}}}. \bibinfo{publisher}{OpenReview.net}.
\newblock


\bibitem[Lu et~al\mbox{.}(2022a)]%
        {0011ZB0L022}
\bibfield{author}{\bibinfo{person}{Cheng Lu}, \bibinfo{person}{Yuhao Zhou}, \bibinfo{person}{Fan Bao}, \bibinfo{person}{Jianfei Chen}, \bibinfo{person}{Chongxuan Li}, {and} \bibinfo{person}{Jun Zhu}.} \bibinfo{year}{2022}\natexlab{a}.
\newblock \showarticletitle{DPM-Solver: {A} Fast {ODE} Solver for Diffusion Probabilistic Model Sampling in Around 10 Steps}. In \bibinfo{booktitle}{\emph{{NeurIPS}}}.
\newblock


\bibitem[Lu et~al\mbox{.}(2022b)]%
        {abs-2211-01095}
\bibfield{author}{\bibinfo{person}{Cheng Lu}, \bibinfo{person}{Yuhao Zhou}, \bibinfo{person}{Fan Bao}, \bibinfo{person}{Jianfei Chen}, \bibinfo{person}{Chongxuan Li}, {and} \bibinfo{person}{Jun Zhu}.} \bibinfo{year}{2022}\natexlab{b}.
\newblock \showarticletitle{DPM-Solver++: Fast Solver for Guided Sampling of Diffusion Probabilistic Models}.
\newblock \bibinfo{journal}{\emph{CoRR}}  \bibinfo{volume}{abs/2211.01095} (\bibinfo{year}{2022}).
\newblock
\urldef\tempurl%
\url{https://doi.org/10.48550/arXiv.2211.01095}
\showDOI{\tempurl}


\bibitem[Mikubill(2023)]%
        {multi-controlnet}
\bibfield{author}{\bibinfo{person}{Lyumin~Zhang Mikubill}.} \bibinfo{year}{2023}\natexlab{}.
\newblock \bibinfo{title}{sd-webui-controlnet}.
\newblock \bibinfo{howpublished}{\url{https://github.com/Mikubill/sd-webui-controlnet}}.
\newblock
\newblock
\shownote{Accessed: DATE 2023-07-01}.


\bibitem[Mou et~al\mbox{.}(2023)]%
        {t2i-adapter}
\bibfield{author}{\bibinfo{person}{Chong Mou}, \bibinfo{person}{Xintao Wang}, \bibinfo{person}{Liangbin Xie}, \bibinfo{person}{Jian Zhang}, \bibinfo{person}{Zhongang Qi}, \bibinfo{person}{Ying Shan}, {and} \bibinfo{person}{Xiaohu Qie}.} \bibinfo{year}{2023}\natexlab{}.
\newblock \showarticletitle{T2I-Adapter: Learning Adapters to Dig out More Controllable Ability for Text-to-Image Diffusion Models}.
\newblock \bibinfo{journal}{\emph{CoRR}}  \bibinfo{volume}{abs/2302.08453} (\bibinfo{year}{2023}).
\newblock
\urldef\tempurl%
\url{https://doi.org/10.48550/ARXIV.2302.08453}
\showDOI{\tempurl}


\bibitem[Parakkat et~al\mbox{.}(2022)]%
        {cgf.14517}
\bibfield{author}{\bibinfo{person}{Amal~Dev Parakkat}, \bibinfo{person}{Pooran Memari}, {and} \bibinfo{person}{Marie-Paule Cani}.} \bibinfo{year}{2022}\natexlab{}.
\newblock \showarticletitle{Delaunay Painting: Perceptual Image Colouring from Raster Contours with Gaps}.
\newblock \bibinfo{journal}{\emph{Computer Graphics Forum}} \bibinfo{volume}{41}, \bibinfo{number}{6} (\bibinfo{year}{2022}), \bibinfo{pages}{166--181}.
\newblock
\urldef\tempurl%
\url{https://doi.org/10.1111/cgf.14517}
\showDOI{\tempurl}


\bibitem[Patashnik et~al\mbox{.}(2021)]%
        {PatashnikWSCL21}
\bibfield{author}{\bibinfo{person}{Or Patashnik}, \bibinfo{person}{Zongze Wu}, \bibinfo{person}{Eli Shechtman}, \bibinfo{person}{Daniel Cohen{-}Or}, {and} \bibinfo{person}{Dani Lischinski}.} \bibinfo{year}{2021}\natexlab{}.
\newblock \showarticletitle{StyleCLIP: Text-Driven Manipulation of StyleGAN Imagery}. In \bibinfo{booktitle}{\emph{{ICCV}}}. \bibinfo{publisher}{{IEEE/CVF}}, \bibinfo{pages}{2065--2074}.
\newblock
\urldef\tempurl%
\url{https://doi.org/10.1109/ICCV48922.2021.00209}
\showDOI{\tempurl}


\bibitem[Podell et~al\mbox{.}(2023)]%
        {sdxl}
\bibfield{author}{\bibinfo{person}{Dustin Podell}, \bibinfo{person}{Zion English}, \bibinfo{person}{Kyle Lacey}, \bibinfo{person}{Andreas Blattmann}, \bibinfo{person}{Tim Dockhorn}, \bibinfo{person}{Jonas M{\"{u}}ller}, \bibinfo{person}{Joe Penna}, {and} \bibinfo{person}{Robin Rombach}.} \bibinfo{year}{2023}\natexlab{}.
\newblock \showarticletitle{{SDXL:} Improving Latent Diffusion Models for High-Resolution Image Synthesis}.
\newblock \bibinfo{journal}{\emph{CoRR}}  \bibinfo{volume}{abs/2307.01952} (\bibinfo{year}{2023}).
\newblock
\urldef\tempurl%
\url{https://doi.org/10.48550/ARXIV.2307.01952}
\showDOI{\tempurl}


\bibitem[Radford et~al\mbox{.}(2021)]%
        {RadfordKHRGASAM21}
\bibfield{author}{\bibinfo{person}{Alec Radford}, \bibinfo{person}{Jong~Wook Kim}, \bibinfo{person}{Chris Hallacy}, \bibinfo{person}{Aditya Ramesh}, \bibinfo{person}{Gabriel Goh}, \bibinfo{person}{Sandhini Agarwal}, \bibinfo{person}{Girish Sastry}, \bibinfo{person}{Amanda Askell}, \bibinfo{person}{Pamela Mishkin}, \bibinfo{person}{Jack Clark}, \bibinfo{person}{Gretchen Krueger}, {and} \bibinfo{person}{Ilya Sutskever}.} \bibinfo{year}{2021}\natexlab{}.
\newblock \showarticletitle{Learning Transferable Visual Models From Natural Language Supervision}. In \bibinfo{booktitle}{\emph{{ICML}}}, Vol.~\bibinfo{volume}{139}. \bibinfo{publisher}{{PMLR}}, \bibinfo{pages}{8748--8763}.
\newblock


\bibitem[Ramesh et~al\mbox{.}(2022)]%
        {abs-2204-06125}
\bibfield{author}{\bibinfo{person}{Aditya Ramesh}, \bibinfo{person}{Prafulla Dhariwal}, \bibinfo{person}{Alex Nichol}, \bibinfo{person}{Casey Chu}, {and} \bibinfo{person}{Mark Chen}.} \bibinfo{year}{2022}\natexlab{}.
\newblock \showarticletitle{Hierarchical Text-Conditional Image Generation with {CLIP} Latents}.
\newblock \bibinfo{journal}{\emph{CoRR}}  \bibinfo{volume}{abs/2204.06125} (\bibinfo{year}{2022}).
\newblock
\urldef\tempurl%
\url{https://doi.org/10.48550/arXiv.2204.06125}
\showDOI{\tempurl}


\bibitem[Rombach et~al\mbox{.}(2022)]%
        {RombachBLEO22}
\bibfield{author}{\bibinfo{person}{Robin Rombach}, \bibinfo{person}{Andreas Blattmann}, \bibinfo{person}{Dominik Lorenz}, \bibinfo{person}{Patrick Esser}, {and} \bibinfo{person}{Bj{\"{o}}rn Ommer}.} \bibinfo{year}{2022}\natexlab{}.
\newblock \showarticletitle{High-Resolution Image Synthesis with Latent Diffusion Models}. In \bibinfo{booktitle}{\emph{{CVPR}}}. \bibinfo{publisher}{{IEEE/CVF}}, \bibinfo{pages}{10674--10685}.
\newblock
\urldef\tempurl%
\url{https://doi.org/10.1109/CVPR52688.2022.01042}
\showDOI{\tempurl}


\bibitem[Ronneberger et~al\mbox{.}(2015)]%
        {RonnebergerFB15}
\bibfield{author}{\bibinfo{person}{Olaf Ronneberger}, \bibinfo{person}{Philipp Fischer}, {and} \bibinfo{person}{Thomas Brox}.} \bibinfo{year}{2015}\natexlab{}.
\newblock \showarticletitle{U-Net: Convolutional Networks for Biomedical Image Segmentation}. In \bibinfo{booktitle}{\emph{{MICCAI}}}, Vol.~\bibinfo{volume}{9351}. \bibinfo{publisher}{Springer}, \bibinfo{pages}{234--241}.
\newblock
\urldef\tempurl%
\url{https://doi.org/10.1007/978-3-319-24574-4\_28}
\showDOI{\tempurl}


\bibitem[rqdwdw(2023)]%
        {negative-xl}
\bibfield{author}{\bibinfo{person}{rqdwdw}.} \bibinfo{year}{2023}\natexlab{}.
\newblock \bibinfo{title}{negativeXL}.
\newblock \bibinfo{howpublished}{\url{https://civitai.com/models/118418/negativexl}}.
\newblock
\newblock
\shownote{Accessed: DATE 2023-02-10}.


\bibitem[Ruiz et~al\mbox{.}(2023)]%
        {ruiz2022dreambooth}
\bibfield{author}{\bibinfo{person}{Nataniel Ruiz}, \bibinfo{person}{Yuanzhen Li}, \bibinfo{person}{Varun Jampani}, \bibinfo{person}{Yael Pritch}, \bibinfo{person}{Michael Rubinstein}, {and} \bibinfo{person}{Kfir Aberman}.} \bibinfo{year}{2023}\natexlab{}.
\newblock \showarticletitle{DreamBooth: Fine Tuning Text-to-Image Diffusion Models for Subject-Driven Generation}. In \bibinfo{booktitle}{\emph{{CVPR}}}. \bibinfo{publisher}{{IEEE/CVF}}, \bibinfo{pages}{22500--22510}.
\newblock
\urldef\tempurl%
\url{https://doi.org/10.1109/CVPR52729.2023.02155}
\showDOI{\tempurl}


\bibitem[runwayml(2024)]%
        {sd1.5-hf}
\bibfield{author}{\bibinfo{person}{runwayml}.} \bibinfo{year}{2024}\natexlab{}.
\newblock \bibinfo{title}{stable-diffusion-v1-5}.
\newblock \bibinfo{howpublished}{\url{https://huggingface.co/runwayml/stable-diffusion-v1-5}}.
\newblock
\newblock
\shownote{Accessed: DATE 2024-01-02}.


\bibitem[Schaefer et~al\mbox{.}(2006)]%
        {SchaeferMW06}
\bibfield{author}{\bibinfo{person}{Scott Schaefer}, \bibinfo{person}{Travis McPhail}, {and} \bibinfo{person}{Joe~D. Warren}.} \bibinfo{year}{2006}\natexlab{}.
\newblock \showarticletitle{Image deformation using moving least squares}.
\newblock \bibinfo{journal}{\emph{{ACM} Trans. Graph.}} \bibinfo{volume}{25}, \bibinfo{number}{3} (\bibinfo{year}{2006}), \bibinfo{pages}{533--540}.
\newblock
\urldef\tempurl%
\url{https://doi.org/10.1145/1141911.1141920}
\showDOI{\tempurl}


\bibitem[Schuhmann et~al\mbox{.}(2022)]%
        {schuhmann2022laionb}
\bibfield{author}{\bibinfo{person}{Christoph Schuhmann}, \bibinfo{person}{Romain Beaumont}, \bibinfo{person}{Richard Vencu}, \bibinfo{person}{Cade~W Gordon}, \bibinfo{person}{Ross Wightman}, \bibinfo{person}{Mehdi Cherti}, \bibinfo{person}{Theo Coombes}, \bibinfo{person}{Aarush Katta}, \bibinfo{person}{Clayton Mullis}, \bibinfo{person}{Mitchell Wortsman}, \bibinfo{person}{Patrick Schramowski}, \bibinfo{person}{Srivatsa~R Kundurthy}, \bibinfo{person}{Katherine Crowson}, \bibinfo{person}{Ludwig Schmidt}, \bibinfo{person}{Robert Kaczmarczyk}, {and} \bibinfo{person}{Jenia Jitsev}.} \bibinfo{year}{2022}\natexlab{}.
\newblock \showarticletitle{{LAION}-5B: An open large-scale dataset for training next generation image-text models}. \bibinfo{howpublished}{\url{https://openreview.net/forum?id=M3Y74vmsMcY}}. In \bibinfo{booktitle}{\emph{Thirty-sixth Conference on Neural Information Processing Systems Datasets and Benchmarks Track}}.
\newblock


\bibitem[Seitzer(2023)]%
        {Seitzer2020FID}
\bibfield{author}{\bibinfo{person}{Maximilian Seitzer}.} \bibinfo{year}{2023}\natexlab{}.
\newblock \bibinfo{title}{{pytorch-fid: FID Score for PyTorch}}.
\newblock \bibinfo{howpublished}{\url{https://github.com/mseitzer/pytorch-fid}}.
\newblock
\newblock
\shownote{Accessed: DATE 2023-05-17}.


\bibitem[Sohl{-}Dickstein et~al\mbox{.}(2015)]%
        {Sohl-DicksteinW15}
\bibfield{author}{\bibinfo{person}{Jascha Sohl{-}Dickstein}, \bibinfo{person}{Eric~A. Weiss}, \bibinfo{person}{Niru Maheswaranathan}, {and} \bibinfo{person}{Surya Ganguli}.} \bibinfo{year}{2015}\natexlab{}.
\newblock \showarticletitle{Deep Unsupervised Learning using Nonequilibrium Thermodynamics}. In \bibinfo{booktitle}{\emph{{ICML}}}, Vol.~\bibinfo{volume}{37}. \bibinfo{publisher}{JMLR.org}, \bibinfo{pages}{2256--2265}.
\newblock


\bibitem[Song et~al\mbox{.}(2021a)]%
        {SongME21}
\bibfield{author}{\bibinfo{person}{Jiaming Song}, \bibinfo{person}{Chenlin Meng}, {and} \bibinfo{person}{Stefano Ermon}.} \bibinfo{year}{2021}\natexlab{a}.
\newblock \showarticletitle{Denoising Diffusion Implicit Models}. In \bibinfo{booktitle}{\emph{{ICLR}}}. \bibinfo{publisher}{OpenReview.net}.
\newblock


\bibitem[Song et~al\mbox{.}(2021b)]%
        {0011SKKEP21}
\bibfield{author}{\bibinfo{person}{Yang Song}, \bibinfo{person}{Jascha Sohl{-}Dickstein}, \bibinfo{person}{Diederik~P. Kingma}, \bibinfo{person}{Abhishek Kumar}, \bibinfo{person}{Stefano Ermon}, {and} \bibinfo{person}{Ben Poole}.} \bibinfo{year}{2021}\natexlab{b}.
\newblock \showarticletitle{Score-Based Generative Modeling through Stochastic Differential Equations}. In \bibinfo{booktitle}{\emph{{ICLR}}}. \bibinfo{publisher}{OpenReview.net}.
\newblock


\bibitem[Stability-AI(2024)]%
        {sdxl-hf}
\bibfield{author}{\bibinfo{person}{Stability-AI}.} \bibinfo{year}{2024}\natexlab{}.
\newblock \bibinfo{title}{stable-diffusion-xl-base-1.0}.
\newblock \bibinfo{howpublished}{\url{https://huggingface.co/stabilityai/stable-diffusion-xl-base-1.0}}.
\newblock
\newblock
\shownote{Accessed: DATE 2024-01-02}.


\bibitem[Sun et~al\mbox{.}(2019)]%
        {SunLWW19}
\bibfield{author}{\bibinfo{person}{Tsai{-}Ho Sun}, \bibinfo{person}{Chien{-}Hsun Lai}, \bibinfo{person}{Sai{-}Keung Wong}, {and} \bibinfo{person}{Yu{-}Shuen Wang}.} \bibinfo{year}{2019}\natexlab{}.
\newblock \showarticletitle{Adversarial Colorization of Icons Based on Contour and Color Conditions}. In \bibinfo{booktitle}{\emph{{ACM} {MM}}}. \bibinfo{publisher}{{ACM}}, \bibinfo{pages}{683--691}.
\newblock
\urldef\tempurl%
\url{https://doi.org/10.1145/3343031.3351041}
\showDOI{\tempurl}


\bibitem[S{\'{y}}kora et~al\mbox{.}(2009)]%
        {SykoraDC09}
\bibfield{author}{\bibinfo{person}{Daniel S{\'{y}}kora}, \bibinfo{person}{John Dingliana}, {and} \bibinfo{person}{Steven Collins}.} \bibinfo{year}{2009}\natexlab{}.
\newblock \showarticletitle{LazyBrush: Flexible Painting Tool for Hand-drawn Cartoons}.
\newblock \bibinfo{journal}{\emph{Comput. Graph. Forum}} \bibinfo{volume}{28}, \bibinfo{number}{2} (\bibinfo{year}{2009}), \bibinfo{pages}{599--608}.
\newblock
\urldef\tempurl%
\url{https://doi.org/10.1111/j.1467-8659.2009.01400.x}
\showDOI{\tempurl}


\bibitem[TencentARC(2024)]%
        {t2i-adapter-code}
\bibfield{author}{\bibinfo{person}{TencentARC}.} \bibinfo{year}{2024}\natexlab{}.
\newblock \bibinfo{title}{Hugging Face/IP-Adapter}.
\newblock \bibinfo{howpublished}{\url{https://github.com/TencentARC/T2I-Adapter/tree/SD}}.
\newblock
\newblock
\shownote{Accessed: DATE 2024-01-02}.


\bibitem[Tumanyan et~al\mbox{.}(2023)]%
        {Tumanyan_2023_CVPR}
\bibfield{author}{\bibinfo{person}{Narek Tumanyan}, \bibinfo{person}{Michal Geyer}, \bibinfo{person}{Shai Bagon}, {and} \bibinfo{person}{Tali Dekel}.} \bibinfo{year}{2023}\natexlab{}.
\newblock \showarticletitle{Plug-and-Play Diffusion Features for Text-Driven Image-to-Image Translation}. In \bibinfo{booktitle}{\emph{{CVPR}}}. \bibinfo{publisher}{{IEEE/CVF}}, \bibinfo{pages}{1921--1930}.
\newblock
\urldef\tempurl%
\url{https://doi.org/10.1109/CVPR52729.2023.00191}
\showDOI{\tempurl}


\bibitem[van~den Oord et~al\mbox{.}(2017)]%
        {OordVK17}
\bibfield{author}{\bibinfo{person}{A{\"{a}}ron van~den Oord}, \bibinfo{person}{Oriol Vinyals}, {and} \bibinfo{person}{Koray Kavukcuoglu}.} \bibinfo{year}{2017}\natexlab{}.
\newblock \showarticletitle{Neural Discrete Representation Learning}. In \bibinfo{booktitle}{\emph{{NeurIPS}}}. \bibinfo{pages}{6306--6315}.
\newblock


\bibitem[Xiang et~al\mbox{.}(2022)]%
        {xiang2022adversarial}
\bibfield{author}{\bibinfo{person}{Xiaoyu Xiang}, \bibinfo{person}{Ding Liu}, \bibinfo{person}{Xiao Yang}, \bibinfo{person}{Yiheng Zhu}, \bibinfo{person}{Xiaohui Shen}, {and} \bibinfo{person}{Jan~P. Allebach}.} \bibinfo{year}{2022}\natexlab{}.
\newblock \showarticletitle{Adversarial Open Domain Adaptation for Sketch-to-Photo Synthesis}. In \bibinfo{booktitle}{\emph{{WACV}}}. \bibinfo{publisher}{{IEEE/CVF}}, \bibinfo{pages}{944--954}.
\newblock
\urldef\tempurl%
\url{https://doi.org/10.1109/WACV51458.2022.00102}
\showDOI{\tempurl}


\bibitem[Yan et~al\mbox{.}(2023)]%
        {yan-cgf}
\bibfield{author}{\bibinfo{person}{Dingkun Yan}, \bibinfo{person}{Ryogo Ito}, \bibinfo{person}{Ryo Moriai}, {and} \bibinfo{person}{Suguru Saito}.} \bibinfo{year}{2023}\natexlab{}.
\newblock \showarticletitle{Two-Step Training: Adjustable Sketch Colourization via Reference Image and Text Tag}.
\newblock \bibinfo{journal}{\emph{Computer Graphics Forum}} (\bibinfo{year}{2023}).
\newblock
\urldef\tempurl%
\url{https://doi.org/10.1111/cgf.14791}
\showDOI{\tempurl}


\bibitem[Ye et~al\mbox{.}(2023)]%
        {ip-adapter}
\bibfield{author}{\bibinfo{person}{Hu Ye}, \bibinfo{person}{Jun Zhang}, \bibinfo{person}{Sibo Liu}, \bibinfo{person}{Xiao Han}, {and} \bibinfo{person}{Wei Yang}.} \bibinfo{year}{2023}\natexlab{}.
\newblock \showarticletitle{IP-Adapter: Text Compatible Image Prompt Adapter for Text-to-Image Diffusion Models}.
\newblock \bibinfo{journal}{\emph{CoRR}}  \bibinfo{volume}{abs/2308.06721} (\bibinfo{year}{2023}).
\newblock
\urldef\tempurl%
\url{https://doi.org/10.48550/ARXIV.2308.06721}
\showDOI{\tempurl}


\bibitem[Yuno779(2023)]%
        {anything}
\bibfield{author}{\bibinfo{person}{Yuno779}.} \bibinfo{year}{2023}\natexlab{}.
\newblock \bibinfo{howpublished}{\url{https://civitai.com/models/9409}}.
\newblock
\newblock
\shownote{Accessed: DATE 2023-06-25}.


\bibitem[Zhang(2017)]%
        {sketchKeras}
\bibfield{author}{\bibinfo{person}{Lvmin Zhang}.} \bibinfo{year}{2017}\natexlab{}.
\newblock \bibinfo{title}{SketchKeras}.
\newblock \bibinfo{howpublished}{\url{https://github.com/lllyasviel/sketchKeras}}.
\newblock


\bibitem[Zhang(2023)]%
        {lvmin-injection}
\bibfield{author}{\bibinfo{person}{Lvmin Zhang}.} \bibinfo{year}{2023}\natexlab{}.
\newblock \bibinfo{title}{How ControlNet-reference works}.
\newblock \bibinfo{howpublished}{\url{https://github.com/Mikubill/sd-webui-controlnet/discussions/1236}}.
\newblock


\bibitem[Zhang(2024)]%
        {controlnet-v11}
\bibfield{author}{\bibinfo{person}{Lvmin Zhang}.} \bibinfo{year}{2024}\natexlab{}.
\newblock \bibinfo{title}{ControlNet-v1-1-nightly}.
\newblock \bibinfo{howpublished}{\url{https://github.com/lllyasviel/ControlNet-v1-1-nightly}}.
\newblock
\newblock
\shownote{Accessed: DATE 2024-01-02}.


\bibitem[Zhang and Agrawala(2023)]%
        {controlnet}
\bibfield{author}{\bibinfo{person}{Lvmin Zhang} {and} \bibinfo{person}{Maneesh Agrawala}.} \bibinfo{year}{2023}\natexlab{}.
\newblock \showarticletitle{Adding Conditional Control to Text-to-Image Diffusion Models}.
\newblock \bibinfo{journal}{\emph{CoRR}}  \bibinfo{volume}{abs/2302.05543} (\bibinfo{year}{2023}).
\newblock
\urldef\tempurl%
\url{https://doi.org/10.48550/arXiv.2302.05543}
\showDOI{\tempurl}


\bibitem[Zhang et~al\mbox{.}(2018)]%
        {ZhangLW0L18}
\bibfield{author}{\bibinfo{person}{Lvmin Zhang}, \bibinfo{person}{Chengze Li}, \bibinfo{person}{Tien{-}Tsin Wong}, \bibinfo{person}{Yi Ji}, {and} \bibinfo{person}{Chunping Liu}.} \bibinfo{year}{2018}\natexlab{}.
\newblock \showarticletitle{Two-stage sketch colorization}.
\newblock \bibinfo{journal}{\emph{{ACM} Trans. Graph.}} \bibinfo{volume}{37}, \bibinfo{number}{6} (\bibinfo{year}{2018}), \bibinfo{pages}{261}.
\newblock
\urldef\tempurl%
\url{https://doi.org/10.1145/3272127.3275090}
\showDOI{\tempurl}


\bibitem[Zhang et~al\mbox{.}(2023b)]%
        {controlnet-iccv}
\bibfield{author}{\bibinfo{person}{Lvmin Zhang}, \bibinfo{person}{Anyi Rao}, {and} \bibinfo{person}{Maneesh Agrawala}.} \bibinfo{year}{2023}\natexlab{b}.
\newblock \showarticletitle{Adding Conditional Control to Text-to-Image Diffusion Models}. In \bibinfo{booktitle}{\emph{{ICCV}}}. \bibinfo{pages}{3836--3847}.
\newblock


\bibitem[Zhang et~al\mbox{.}(2016)]%
        {ZhangIE16}
\bibfield{author}{\bibinfo{person}{Richard Zhang}, \bibinfo{person}{Phillip Isola}, {and} \bibinfo{person}{Alexei~A. Efros}.} \bibinfo{year}{2016}\natexlab{}.
\newblock \showarticletitle{Colorful Image Colorization}. In \bibinfo{booktitle}{\emph{{ECCV}}}, Vol.~\bibinfo{volume}{9907}. \bibinfo{publisher}{Springer}, \bibinfo{pages}{649--666}.
\newblock
\urldef\tempurl%
\url{https://doi.org/10.1007/978-3-319-46487-9\_40}
\showDOI{\tempurl}


\bibitem[Zhang et~al\mbox{.}(2017)]%
        {ZhangZIGLYE17}
\bibfield{author}{\bibinfo{person}{Richard Zhang}, \bibinfo{person}{Jun{-}Yan Zhu}, \bibinfo{person}{Phillip Isola}, \bibinfo{person}{Xinyang Geng}, \bibinfo{person}{Angela~S. Lin}, \bibinfo{person}{Tianhe Yu}, {and} \bibinfo{person}{Alexei~A. Efros}.} \bibinfo{year}{2017}\natexlab{}.
\newblock \showarticletitle{Real-time user-guided image colorization with learned deep priors}.
\newblock \bibinfo{journal}{\emph{{ACM} Trans. Graph.}} \bibinfo{volume}{36}, \bibinfo{number}{4} (\bibinfo{year}{2017}), \bibinfo{pages}{119:1--119:11}.
\newblock
\urldef\tempurl%
\url{https://doi.org/10.1145/3072959.3073703}
\showDOI{\tempurl}


\bibitem[Zhang et~al\mbox{.}(2023a)]%
        {zhang2023prospect}
\bibfield{author}{\bibinfo{person}{Yuxin Zhang}, \bibinfo{person}{Weiming Dong}, \bibinfo{person}{Fan Tang}, \bibinfo{person}{Nisha Huang}, \bibinfo{person}{Haibin Huang}, \bibinfo{person}{Chongyang Ma}, \bibinfo{person}{Tong-Yee Lee}, \bibinfo{person}{Oliver Deussen}, {and} \bibinfo{person}{Changsheng Xu}.} \bibinfo{year}{2023}\natexlab{a}.
\newblock \showarticletitle{ProSpect: Prompt Spectrum for Attribute-Aware Personalization of Diffusion Models}.
\newblock \bibinfo{journal}{\emph{{ACM} Trans. Graph.}} \bibinfo{volume}{42}, \bibinfo{number}{6} (\bibinfo{year}{2023}), \bibinfo{pages}{244:1--244:14}.
\newblock


\bibitem[Zhang et~al\mbox{.}(2023c)]%
        {abs-2305-18729}
\bibfield{author}{\bibinfo{person}{Yuechen Zhang}, \bibinfo{person}{Jinbo Xing}, \bibinfo{person}{Eric Lo}, {and} \bibinfo{person}{Jiaya Jia}.} \bibinfo{year}{2023}\natexlab{c}.
\newblock \showarticletitle{Real-World Image Variation by Aligning Diffusion Inversion Chain}.
\newblock \bibinfo{journal}{\emph{CoRR}}  \bibinfo{volume}{abs/2305.18729} (\bibinfo{year}{2023}).
\newblock
\urldef\tempurl%
\url{https://doi.org/10.48550/arXiv.2305.18729}
\showDOI{\tempurl}


\bibitem[Zhu et~al\mbox{.}(2017)]%
        {ZhuPIE17}
\bibfield{author}{\bibinfo{person}{Jun{-}Yan Zhu}, \bibinfo{person}{Taesung Park}, \bibinfo{person}{Phillip Isola}, {and} \bibinfo{person}{Alexei~A. Efros}.} \bibinfo{year}{2017}\natexlab{}.
\newblock \showarticletitle{Unpaired Image-to-Image Translation Using Cycle-Consistent Adversarial Networks}. In \bibinfo{booktitle}{\emph{{ICCV}}}. \bibinfo{publisher}{{IEEE/CVF}}, \bibinfo{pages}{2242--2251}.
\newblock
\urldef\tempurl%
\url{https://doi.org/10.1109/ICCV.2017.244}
\showDOI{\tempurl}


\bibitem[Zou et~al\mbox{.}(2019)]%
        {zouSA2019sketchcolorization}
\bibfield{author}{\bibinfo{person}{Changqing Zou}, \bibinfo{person}{Haoran Mo}, \bibinfo{person}{Chengying Gao}, \bibinfo{person}{Ruofei Du}, {and} \bibinfo{person}{Hongbo Fu}.} \bibinfo{year}{2019}\natexlab{}.
\newblock \showarticletitle{Language-Based Colorization of Scene Sketches}.
\newblock \bibinfo{journal}{\emph{{ACM} Trans. Graph.}} \bibinfo{volume}{38}, \bibinfo{number}{6} (\bibinfo{year}{2019}).
\newblock
\urldef\tempurl%
\url{https://doi.org/10.1145/3355089.3356561}
\showDOI{\tempurl}


\end{thebibliography}

\end{document}

\title{Supplementary materials}
\section{Improvement on Generation}
\begin{figure}[t]
	\centering
	\includegraphics[width=1\linewidth]{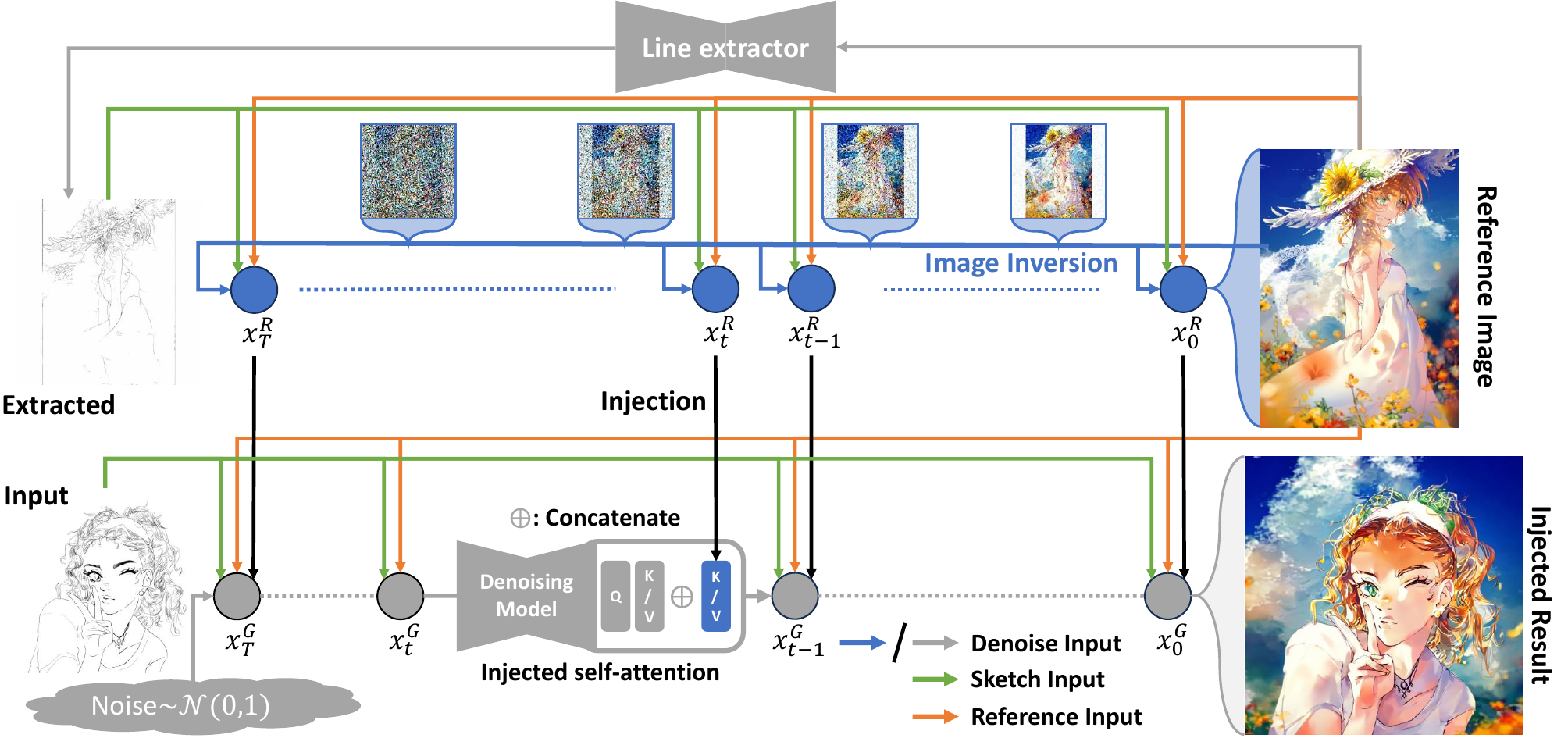}
	\caption{Illustration of our attention injection. We adopt \cite{xiang2022adversarial} as our default line extractor.}
	\label{injection}
\end{figure}
We introduce several important suggestions that can further improve the generation performance.\\

\noindent\textbf{Resolution.} Increasing the image resolution significantly improves reference-based sketch colorization. Sketch images in higher resolution provide detailed strokes and richer semantic information. Experimentally, optimal inference results often manifest at 1.5x the training resolution, e.g., training at $512^{2}$ and inferring at $768^{2}$. Real color images created by experienced artists contain numerous visual attributes that are difficult to transfer fully. However, reference-based models always manage to generate all these attributes in the sketch image, leading to overly saturated colors. Utilizing a larger resolution during inference can effectively moderate these reference features, yielding more appealing results.\\

\noindent\textbf{Attention injection and AdaIN.} Our implementation of attention injection and AdaIN is similar to that of \textit{ControlNet-reference} \cite{multi-controlnet,lvmin-injection}, and both techniques could be adopted to improve our generated results.  Here, we briefly introduce how the attention injection is adapted to our reference-based colorization models. As illustrated in Figure \ref{injection}, we utilize a sketch extracted from the reference image as the sketch input for the inversion $\bm{x}^{R}$ chain. Given the intermediate hidden states $\bm{h}^{R}$ obtained from the $\bm{x}^{R}$ chain, and $\bm{h}^{G}$ from the generation $\bm{x}^{G}$ chain, we concatenate them as $\bm{h}_{c}^{G}$ for computing $K$ and $V$ in the self-attention modules, calculated as:"
\begin{equation}
	\begin{aligned}
	Q=W_{q}\cdot\bm{h}^{G},\enspace&K=W_{k}\cdot\bm{h}^{G}_{c},\enspace V=W_{v}\cdot\bm{h}^{G}_{c}, where \\
	&\bm{h}_{c}^{G}=\bm{h}^{R}\oplus\bm{h}^{G}
	\end{aligned}
\end{equation}
where, $W_{q},W_{k}$ and $W_{v}$ denote the weight matrix for $Q,K$ and $V$, respectively.\\
\bibliographystyle{ACM-Reference-Format} 
\bibliography{sample-base}